\documentclass[jair,twoside,11pt,theapa]{article}
\usepackage{jair, theapa, rawfonts}
\usepackage{graphicx}
\usepackage{amsmath}
\usepackage{bm}
\usepackage{caption}
\usepackage{subcaption}

\usepackage{amsmath, amsfonts, algpseudocode, algorithm}
\usepackage[all]{xy}
\usepackage{url}

\newenvironment{compress}{\parindent1.5ex\hangindent=1.5ex\baselineskip=0.1\baselineskip\scriptsize\hrule\vspace{1mm}}{\par\vskip -1mm \hrule}
\newcommand{\bb}[1]{\mbox{}\hspace{-1ex}\textbf{#1}\newline}
\newcommand{\p}[1]{\textsl{#1}}
\newcommand{\eat}[1]{}

\ShortHeadings{Neural-Symbolic Learning and Reasoning}
{Besold et al.}
\firstpageno{1}

\begin{document}

\title{Neural-Symbolic Learning and Reasoning:\\A Survey and Interpretation}

\author{\name Tarek R. Besold \email tarek-r.besold@city.ac.uk \\
       \addr Department of Computer Science, City, University of London
       \AND
       \name Artur d'Avila Garcez \email a.garcez@city.ac.uk \\
       \addr Department of Computer Science, City, University London
       \AND
       \name Sebastian Bader \email sebastian.bader@uni-rostock.de \\
       \addr Department of Computer Science, University of Rostock
       \AND
       \name Howard Bowman \email h.bowman@kent.ac.uk \\
       \addr School of Computing, University of Kent
       \AND
       \name Pedro Domingos \email pedrod@cs.washington.edu \\
       \addr Department of Computer Science \& Engineering, University of Washington
       \AND
       \name Pascal Hitzler \email pascal.hitzler@wright.edu \\
       \addr Department of Computer Science \& Engineering, Wright State University
       \AND
       \name Kai-Uwe K\"{u}hnberger \email kkuehnbe@uni-osnabrueck.de \\
       \addr Institute of Cognitive Science, University of Osnabr\"{u}ck
       \AND
       \name Luis C. Lamb \email luislamb@acm.org \\
       \addr Instituto de Informatica, Universidade Federal do Rio Grande do Sul
       \AND
       \name Daniel Lowd \email lowd@cs.uoregon.edu \\
       \addr Department of Computer and Information Science, University of Oregon
       \AND
       \name Priscila Machado Vieira Lima \email priscilamvl@gmail.com \\
       \addr NCE, Universidade Federal do Rio de Janeiro
       \AND
       \name Leo de Penning \email leo.depenning@illuminoo.com \\
       \addr Illuminoo B.V.
       \AND
       \name Gadi Pinkas \email pinkas@gmail.com \\
       \addr Center for Academic Studies and Gonda Brain Research Center, Bar-Ilan University, Israel
       \AND
       \name Hoifung Poon \email hoifung@microsoft.com \\
       \addr Microsoft Research
       \AND
       \name Gerson Zaverucha \email gerson@cos.ufrj.br \\
       \addr COPPE, Universidade Federal do Rio de Janeiro
}

% For research notes, remove the comment character in the line below.
% \researchnote

\maketitle

\begin{abstract}
The study and understanding of human behaviour is relevant to computer science, artificial intelligence, neural computation, cognitive science, philosophy, psychology, and several other areas. Presupposing cognition as basis of behaviour, among the most prominent tools in the modelling of behaviour are computational-logic systems, connectionist models of cognition, and models of uncertainty. Recent studies in cognitive science, artificial intelligence, and psychology have produced a number of cognitive models of reasoning, learning, and language that are underpinned by computation. In addition, efforts in computer science research have led to the development of cognitive computational systems integrating machine learning and automated reasoning. Such systems have shown promise in a range of applications, including computational biology, fault diagnosis, training and assessment in simulators, and software verification. This joint survey reviews the personal ideas and views of several researchers on neural-symbolic learning and reasoning. The article is organised in three parts: Firstly, we frame the scope and goals of neural-symbolic computation and have a look at the theoretical foundations. We then proceed to describe the realisations of neural-symbolic computation, systems, and applications. Finally we present the challenges facing the area and avenues for further research.
\end{abstract}

\section{Overview} \label{overview}
The study of human behaviour is an important part of computer science, artificial intelligence (AI), neural computation, cognitive science, philosophy, psychology, and other areas. Presupposing that behaviour is generally determined and guided by cognition and mental processing, among the most prominent tools in the modelling of behaviour are computational-logic systems mostly addressing high-level reasoning and thought processes (classical logic, nonmonotonic logic, modal and temporal logic), connectionist models of cognition and the brain mostly addressing lower-level dynamics and emergent processes (feedforward and recurrent networks, symmetric and deep networks, self-organising networks), and models of uncertainty addressing the often vague or probabilistic nature of many aspects of cognitive processing (Bayesian networks, Markov decision processes, Markov logic networks, probabilistic inductive logic programs).

Recent studies in cognitive science, artificial intelligence, and psychology have produced a number of cognitive models of reasoning, learning, and language that are underpinned by computation \cite{pinker_2008,shastri_2007,sun_2009}. In addition, recent efforts in computer science have led to the development of cognitive computational systems integrating machine learning and automated reasoning \cite{garcez_2002,garcez_2009,valiant_2000}. Such systems have shown promise in a range of applications, including fault diagnosis, computational biology, training and assessment in simulators, and software verification \cite{depenning_2010,depenning_2011}. 

Forestalling the presentation of the theoretical foundations in Section~\ref{prolegomena}, the intuition motivating neural-symbolic integration as an active field of research is the following: In neural computing, it is assumed that the mind is an emergent property of the brain, and that computational cognitive modelling can lead to valid theories of cognition and offer an understanding of certain cognitive processes \cite{sun_2009}. From this it is in turn assumed that connectionism should be able to offer an appropriate representational language for artificial intelligence as well. In particular, a connectionist computational theory of the mind should be able to replicate the parallelism and kinds of adaptive learning processes seen in neural networks, which are generally accepted as responsible for the necessary robustness and ultimate effectiveness of the system in dealing with commonsense knowledge. As a result, a purely symbolic approach would not be sufficient, as argued by Valiant in \cite{valiant_2008}.

On the other hand, logic is firmly established as a fundamental tool in the modelling of thought and behaviour \cite{kowalski_2011,pereira_2012} and by many has been viewed generally as the ``calculus of computer science''. In this context, often also nonclassical logics play an important role: Temporal logic, for instance, has had significant impact in both academia and industry \cite{pnueli_1977}, and different modal logics have become a lingua franca for, among others, the specification and analysis of knowledge and communication in multi-agent and distributed systems \cite{fagin_1995}. Research on practical reasoning in AI has been dominated by nonmonotonic formalisms. Intuitionistic logic can provide an adequate logical foundation for several core areas of theoretical computer science, including type theory and functional programming \cite{vandalen_2002}. Finally, description logics---which are similar to Kripke models---have been instrumental in the study of the semantic web \cite{baader_2003}.

However, when building models that combine learning and reasoning, one has to conciliate the methodologies of distinct areas---namely predominantly statistics and logic---in order to combine the respective advantages and circumvent the shortcomings and limitations. For instance, the methodology of neural-symbolic systems aims to transfer principles and mechanisms between (often nonclassical) logic-based computation and neural computation. In particular, it considers how principles of symbolic computation can be implemented by connectionist mechanisms and how subsymbolic computation can be described and analysed in logical terms. Here, connectionism provides the hardware upon which different levels of abstraction can be built according to the needs of the application. This methodology---looking at principles, mechanisms, and applications---has proven a fruitful way of progressing the research in the area of neural-symbolic integration for more than two decades now as evidenced by, for instance, the results summarised in the overview by \cite{bader_survey_2005}, collected in the books by \cite{hammer_2007} and \cite{garcez_2009}, and reported in the present survey.

For example in \cite{garcez_2009}, the described approach has led to a prototypical connectionist system for nonclassical reasoning in an attempt to find an adequate balance between complexity and expressiveness. In this framework---known as a neural-symbolic system---artificial neural networks (ANNs) provide the machinery for parallel computation and robust learning, while logic provides the necessary explanation for the network models, facilitating the necessary interaction with the world and other systems. In the integrated model, no conflict arises between a continuous and a discrete component of the system. Instead, a tightly-coupled hybrid system exists that is continuous by nature (the ANN), but that has a clear discrete interpretation (its logic) at various levels of abstraction.

From a more practical perspective, rational agents are often conceptualised as performing concept acquisition (generally unsupervised and statistical) and concept manipulation (generally supervised and symbolic) as part of a permanent cycle of perception and action. The question of how to reconcile the statistical nature of learning with the logical nature of reasoning, aiming to build such robust computational models integrating concept acquisition and manipulation, has been identified as a key research challenge and fundamental problem in computer science \cite{valiant_2003}. Against this backdrop we see neural-symbolic integration as a way of addressing stated challenge through the mechanisms of knowledge translation and knowledge extraction between symbolic logic systems and subsymbolic networks.

There are also important applications of neural-symbolic integration with high relevance for industry applications. The merging of theory (known as background knowledge in machine learning) and data learning (i.e. learning from examples) in ANNs has been shown more effective than purely symbolic or purely connectionist systems, especially in the case of real-world, noisy, unstructured data \cite{depenning_2010,depenning_2011,towell_1994}. Here, successfully addressed application scenarios include business process modelling, service-oriented computing (trust management and fraud prevention in e-commerce), synchronisation and coordination in large multi-agent systems, and multimodal processing and integration.

In multimodal processing, for example, there are several forms of reasoning: a scene classification can be achieved by the well-trained network giving an immediate answer following a number of assumptions. A change in the scene, however, may require more specific temporal, nonmonotonic reasoning and learning from data (based, for example, on the amount of change in the scene). Some assumptions may need to be revised, information from an image annotation may provide a different context, abduction and similarity reasoning by intersecting network ensembles may be needed, probability distributions may have to be reasoned about, and so on. The integrated system will need to respond quickly, revise its answers in the presence of new information, and control the inevitable accumulation of errors derived from real-world data (i.e. prove its robustness). This provides an excellent opportunity for the application of neural-symbolic systems.

The remainder of the paper is structured as follows. In Section~\ref{prolegomena}, we describe the principles of neural-symbolic computation, revisit the theoretical underpinnings of the endeavour, and give a prototypical example of how one can combine learning and reasoning in an integrated fashion. In Section~\ref{learning_and_reasoning} we illustrate the application of the methodology using {\sc NSCA}, a neural-symbolic agent endowed with learning and reasoning capabilities, as a first detailed example. Section~\ref{mental_models} relates concepts underpinning the theories of mind in psychology and cognitive science and their counterparts in neural-symbolic computation, before Section~\ref{logical_neurons} outlines work addressing the binding problem (introduced in the preceding section), thus enabling first-order inference computed by neural-symbolic systems. Section~\ref{connectionist_first-order_logic} then highlights the more technical foundations of first-order (predicate) logic learning in connectionist systems, followed in Section~\ref{markov_logic} by an introduction to Markov logic and corresponding networks as a combination between logic and graphical models. This leads into the conceptual considerations given in Section \ref{hlai} which relate neural-symbolic computation to recent developments in AI and the reinvigorated interest in (re-)creating human-level capacities with artificial systems, notably recently for instance in the area of language modelling and processing. Finally, Section~\ref{recent_developments} summarises selected currently ongoing and widely recognised approaches to solving core questions of neural-symbolic integration arising from neighbouring research efforts and disciplines, before Sections~\ref{future_directions} and \ref{conclusion} present suggested directions for further research and conclude the survey.

\section{Prolegomena of Neural-Symbolic Computation} \label{prolegomena}
The goals of neural-symbolic computation are to provide a coherent, unifying view for logic and connectionism, to contribute to the modelling and understanding of cognition and, thereby, behaviour, and to produce better computational tools for integrated machine learning and reasoning. To this end, logic and network models are studied together as integrated models of computation. Typically, translation algorithms from a symbolic to a connectionist representation and vice-versa are employed to provide either (i) a neural implementation of a logic, (ii) a logical characterisation of a neural system, or (iii) a hybrid learning system that brings together features from connectionism and symbolic artificial intelligence.

From a theoretical perspective, these efforts appear well-founded. According to our current knowledge and understanding, both symbolic/cognitive and sub-symbolic/neural models---especially when focusing on physically-realisable and implementable systems (i.e. physical finite state machines) rather than strictly abstract models of computation, together with the resulting physical and conceptual limitations---seem formally equivalent in a very basic sense:  notwithstanding partially differing theoretical arguments such as given by \cite{tabor_2009}, both paradigms are considered in practice equivalent concerning computability \cite{siegelmann_1999}. Also from a tractability perspective, for instance in \cite{vanrooij_2008}, equivalence in practice with respect to classical dimensions of analysis (i.e. interchangeability except for a polynomial overhead) has been established, complementing and supporting the prior theoretical suggestion of equivalence in the widely accepted Invariance Thesis of \cite{vanemdeboas_1990}. Finally, \cite{leitgeb_2005} provided an \emph{in principle} existence result, showing that there is no substantial difference in representational or problem-solving power between dynamical systems with distributed representations and symbolic systems with non-monotonic reasoning capabilities.

But while these findings provide a solid foundation for attempts at closing the gap between connectionism and logic, many questions nonetheless remain unanswered especially when crossing over from the realm of theoretical research to implementation and application, among others switching from compositional symbols denoting an idealised reality to virtually real-valued vectors obtained from sensors in the real world: Although introducing basic connections and mutual dependencies between the symbolic and the subsymbolic paradigm, the levels of analysis are quite coarse and almost all results are only existential in character. For instance, while establishing the \emph{in principle} equivalence described above, \cite{leitgeb_2005} does not provide constructive methods for how to actually obtain the corresponding symbolic counterpart to a sub-symbolic model and vice versa. 

Still, over the last decades several attempts have been made at developing a general neural-symbolic framework, usually trying to apply the most popular methods of their respective time---such as currently modular deep networks. Growing attention has been given recently to deep networks where it is hoped that high-level abstract representations will emerge from low-level unprocessed data \cite{hinton_2006}. Most modern neural-symbolic systems use feedforward and recurrent networks, but seminal work in the area used symmetric networks \cite{pinkas_1991} of the kind applied in deep learning, and recent work starts to address real applications of symmetric neural-symbolic networks \cite{depenning_2010}. There, in general each level of a neural-symbolic system represents the knowledge evolution of multiple agents over time. Each agent is represented by a network in this level encoding commonsense (nonmonotonic) knowledge and preferences. The networks/agents at different levels can be combined upwards to represent relational knowledge and downwards to create specialisations, following what is known as a network-fibring methodology \cite{garcez_2004}.

Fibring---which will continue to serve as general example throughout the remainder of this section---offers a principled way of combining networks and can be seen as one of the general methodologies of neural-symbolic integration. The main idea of network fibring is simple: fibred networks may be composed of interconnected neurones, as usual, but also of other networks, forming a recursive structure. A fibring function defines how this network architecture behaves; it defines how the networks should relate to each other. Typically, the fibring function will allow the activation of neurones in one network $A$ to influence the change of weights in another network $B$. Intuitively, this may be seen as training network $B$ at the same time that network $A$ is running. Albeit being a combination of simple and standard ANNs, fibred networks can approximate any polynomial function in an unbounded domain, thus being more expressive than standard feedforward networks. 

Fibring is just one example of how principles from symbolic computation (in this case, recursion), can be used by connectionism to advance the research in this area. In the remainder of this section, we discuss in more detail---also in way of a manifesto summarising our view(s) on neural-symbolic integration---the principles, mechanisms and applications that drive the research in neural-symbolic integration.

\subsection{Principles of  Neural-Symbolic Integration} 
From the beginning of connectionism \cite{mcculloch_1943}---arguably the first neural-symbolic system for Boolean logic---most neural-symbolic systems have focused on representing, computing, and learning languages other than classical propositional logic \cite{browne_2001,cloete_2000,garcez_2002,garcez_2009,hoelldobler_1994,shastri_2007}, with much effort being devoted to representing fragments of classical first-order logic. In \cite{garcez_2009}, a new approach to knowledge representation and reasoning has been proposed, establishing connectionist nonclassical logic (including connectionist modal, intuitionistic, temporal, nonmonotonic, epistemic and relational logic). More recently, it has been shown that argumentation frameworks, abductive reasoning, and normative multi-agent systems can also be represented by the same network framework. This is encouraging to the extent that a variety of forms of reasoning can be realised by the same, simple network structure that specialises in different ways.
%Still, from a general perspective the nonclassical approach to neural-symbolic integration hitherto finds its most adequate language for the purpose of integration in propositional modal logic. The latter seems capable of striking the right balance between expressiveness and complexity, it is decidable, strictly more expressive than propositional logic, and, in fact, equivalent to the two-variable fragment of first-order logic. In contrast, the first-order approach to neuro-symbolism has to deal with the problem of variable manipulation and binding. Nevertheless, this is an important problem \cite{bader_2007}, and new insight from the area of lifted message passing may be useful in tackling it \cite{kersting_2010}.

A key characteristic of many neural-symbolic systems is modularity. One way of building neural-symbolic networks is through the careful engineering of network ensembles, where modularity then serves an important role for comprehensibility and maintenance. Each network in the ensemble can be responsible for a specific task or logic, with the overall model being potentially very expressive despite its relatively simple components. Still, although being fairly common, modularity is not a strict necessity: alternatively, as described in Section \ref{logical_neurons}, one can start with an unstructured network (i.e. not modular) and let weight changes shape its ability to process symbolic or subsymbolic representations.

Another common organisational property of neural-symbolic networks is---similar to deep networks---their generally hierarchical organisation. The lowest-level network takes raw data as input and produces a model of the dataset. The next-level network takes the first network's output as its input and produces some higher-level representation of the information in the data. The next-level network then further increases the level of abstraction of the model, and so on, until some high-level representation can be learned. The idea is that such networks might be trained independently, possibly also combining unsupervised and supervised learning at different levels of the hierarchy. The resulting parallel model of computation can be very powerful as it offers the extra expressiveness required by complex applications at comparatively low computational costs \cite{garcez_2009}.

\subsection{Mechanisms of Neural-Symbolic Integration} 
We subscribe to the view that representation initially precedes learning. Neural-symbolic networks can represent a range of expressive logics and implement certain important principles of symbolic computation. However, neural-symbolic computation is not just about representation. The mechanisms of propagation of activation and other message passing methods, gradient-descent and other learning algorithms, reasoning about uncertainty, massive parallelism, fault tolerance, etc. are a crucial part of neural-symbolic integration. Put simply, neural-symbolic networks are efficient computational models, not representational tools. It is the mechanisms in place in the form of efficient algorithms that enable the computational feasibility of neural-symbolic systems.

Returning to the example of fibring, also fibred networks are computational models, not just graphical models or mathematical abstractions like graphs or networks generally. The neural-symbolic networks can be mapped directly onto hardware which promises, for instance, an implementation in a Very-large-scale integration (VLSI) chip to be straightforward and cost effective. The main architectural constraint, which here is brain-inspired, is that neural-symbolic systems should replicate and specialise simple neuronal structures to which a single algorithm can be applied efficiently at different levels of abstraction, with the resulting system being capable of exhibiting emergent behaviour.

It is precisely this emergence as common characterising feature of connectionist approaches which prompts the need for mechanisms of knowledge extraction. The cycle of neural-symbolic integration therefore includes (i) translation of symbolic (background) knowledge into the network, (ii) learning of additional knowledge from examples (and generalisation) by the network, (iii) executing the network (i.e. reasoning), and (iv) symbolic knowledge extraction from the network. Extraction provides explanation, and facilitates maintenance and incremental or transfer learning.

In a general neural-symbolic system, a network ensemble $A$ (representing, for example, a temporal theory) can be combined with another network ensemble $B$ (representing, for example, an agent's epistemic state). Again using fibring as exemplary  mechanism, meta-level knowledge in one network can be integrated with object-level knowledge in another network. For example, one may reason (in the meta-level) about the actions that are needed to be taken (in the object-level) to solve inconsistencies in a database. Relational knowledge can also be represented in the same way, with relations between concepts encoded in distinct (object-level) networks potentially being represented and learned through a meta-level network. More concretely, if two networks denote concepts $P(X,Y)$ and $Q(Z)$ containing variables $X$, $Y$ and $Z$, respectively, a meta-level network can be used to map a representation of $P$ and $Q$ onto a new concept, say $R(X,Y,Z)$, such that, for example, the relation $P(X,Y) \wedge Q(Z) \rightarrow R(X,Y,Z)$ is valid \cite{garcez_2009}.

\begin{figure}
\begin{center}
\includegraphics[width = 0.65\textwidth]{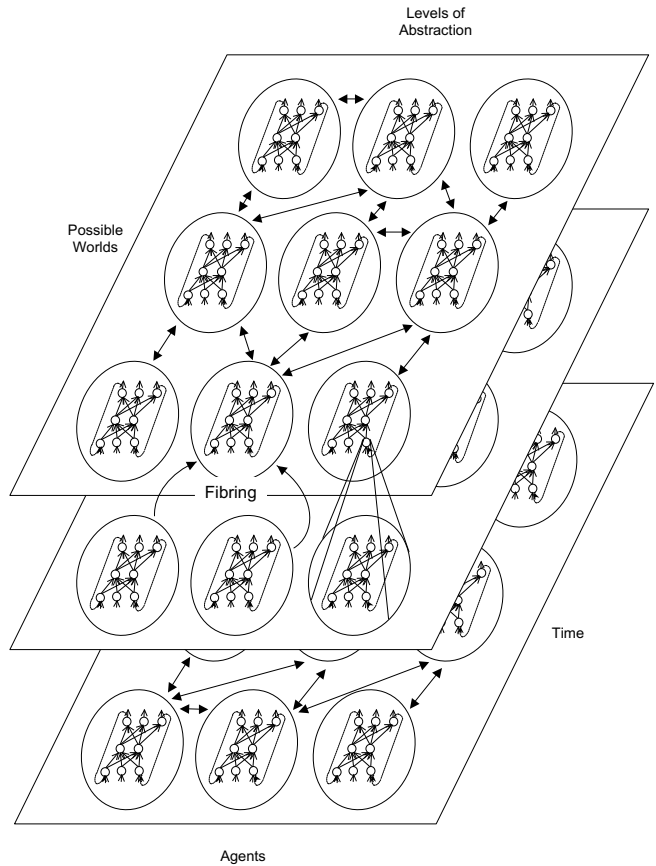}
\caption{General conceptual overview of a neural-symbolic system \cite{garcez_2009}.}\label{neural-symbolic_system}
\end{center}
\end{figure}

Figure~\ref{neural-symbolic_system} illustrates a prototypical neural-symbolic system. The model, in its most general form, allows a number of network ensembles to be combined at different levels of abstraction through, for instance, fibring. In the figure, each level is represented by a network ensemble in a horizontal plane, while network fibring takes place vertically among networks at different ensembles. Specialisation occurs downwards when a neurone is fibred onto a network. Relational knowledge is represented upwards when multiple networks are combined onto a meta-level network. Knowledge evolution through time occurs at each level, as do alternative outcomes, and nonmonotonic and epistemic reasoning for multiple, interacting agents. Modular learning takes place inside each network, but is also applicable across multiple networks in the ensemble. The same brain-inspired structure is replicated throughout the model so that a single algorithm is applied at each level and across levels.

\subsection{Applications of Neural-Symbolic Integration} 
As so often, it will ultimately be through successful implementations and real-world applications that the usefulness and importance of neural-symbolic integration will be widely acknowledged. Practical applications are a crucial ingredient and have been a permanent feature of the research on neural-symbolic computation. On the theory side---in parallel to the quest for understanding the interplay and integration between connectionism and symbolism as the basis of cognition and intelligence---we are interested in finding the limits of representation and developing better machine learning methods which may open up new applications for consideration. In practice, already at the current state of the research, real applications are possible in areas with societal relevance and/or potentially high economic impact, like bioinformatics, semantic web, fault diagnosis, robotics, software systems verification, business process modelling, fraud prevention, multimodal processing, noisy text analytics, and training and assessment in simulators.

All these different application areas have something in common: a computational system is required that is capable of learning from experience and which may reason about what has been learned \cite{browne_2001,valiant_2003}. For this learning-reasoning process to be successful, the system must be robust (in a way so that the accumulation of errors resulting from the intrinsic uncertainty associated with the problem domain can be controlled \cite{valiant_2003}). The method used by the research on neural-symbolic integration to enable some of the above applications has been (i) to apply translation algorithms between logic and networks (making use of the associated equivalence proofs), (ii) to study the systems empirically through case studies (following the practical motivation from statistical machine learning and neural computation), and (iii) to focus on the needs of the application (noting that some potentially interesting applications require just even rudimentary logical representations, e.g. \cite{becker_2009}). An example of a neural-symbolic system that is already providing a contribution of this type to problems in bioinformatics and fault diagnosis is the Connectionist Inductive Learning and Logic Programming ({\sc CILP}) System \cite{garcez_1999,garcez_2002}. Similarly, the {\sc NSCA} system discussed in Section \ref{learning_and_reasoning} has successfully been applied, for instance, to behaviour modelling in training simulators.

\subsection{Neural-Symbolic Integration in a Nutshell}
In summary, neural-symbolic computation encompasses the high-level integration of cognitive abilities (including induction, deduction, and abduction) and the study of how brains make mental models \cite{thagard_2011}, among others also covering the modelling of emotions and attention/utility. At the computational level, it addresses the study of the integration of logic, probabilities, and learning, and the development of new models of computation combining robust learning and efficient reasoning. With regard to applications, successes have been achieved in various domains, including simulation, bioinformatics, fault diagnosis, software engineering, model checking, visual information processing, and fraud prevention. 

In the following section, in order to further familiarise the reader with the overall mindset we will have a detailed look at one such system---namely a neural-symbolic cognitive agent architecture called {\sc NSCA}---aligning the presentation with the three categories principles, methods, and applications.

\section{{\sc NSCA} as Application Example for Neural-Symbolic Computing} \label{learning_and_reasoning}
It is commonly agreed that an effective integration of automated learning and cognitive reasoning in real-world applications is a difficult task \cite{valiant_2003}. Usually, most applications deal with large amounts of data observed in the real-world containing errors, missing values, and inconsistencies. Even in controlled environments, like training simulators, integrated learning and reasoning is not very successful \cite{sandercock_2004,heuvelink_thesis_2009}. Although the use of simulated environments simplifies the data and knowledge acquisition, it is still very difficult to construct a cognitive model of an (intelligent) agent that is able to deal with the many complex relations in the observed data. For example, when it comes to the assessment and training of high-level complex cognitive abilities (e.g. leadership, tactical manoeuvring, safe driving, etc.) training is still guided or done by human experts \cite{vandenbosch_2004}. The reason is that expert behaviour on high-level cognition is too complex to model, elicit, and represent in an automated system. Among others, there can be many temporal relations between low- and high-order aspects of a training task, human behaviour is often non-deterministic and subjective (i.e. biased by personal experience and other factors like stress or fatigue), and what is known is often described vaguely and limited to explicit (i.e. ``explainable'') behaviour. 

Several attempts have been made to tackle these problems. For instance \cite{fernlund_2006} describes a number of systems that use machine learning to learn the complex relations from observations of experts and trainees during task execution. Although these systems are successful at learning and generalisation, they lack the expressive power of symbolic systems and are therefore difficult to interpret and validate \cite{smith_2006}. Alternatively, one could add probabilistic reasoning to logic-based systems \cite{heuvelink_thesis_2009}. These systems perform better in expressing their internal knowledge as they use explicit symbolic representations and are able to deal with many of the most common types of inconsistencies in the data by reasoning with probabilities. Unfortunately, when it comes to knowledge representation and modelling, these systems still require either statistical analyses of large amounts of data or knowledge representation by hand. Therefore, both approaches are time expensive and are not appropriate for use in real-time applications, which demand online learning and reasoning.

\subsection{Principles of Neural-Symbolic Integration Exemplified} The construction of effective cognitive agent models is a long-standing research endeavour in artificial intelligence, cognitive science, and multi-agent systems \cite{valiant_2003,woolridge_2009}. One of the main challenges toward achieving such models is the provision of integrated cognitive abilities, such as learning, reasoning, and knowledge representation. Neural-symbolic systems seek to do so within the neural computation paradigm, integrating inductive learning and deductive reasoning (cf. \cite{garcez_2009,lehmann_2010} for examples). In such models, ANNs are used to learn and reason about (an agent's) knowledge about the world, represented by symbolic logic. In order to do so, algorithms map logical theories (or knowledge about the world) $T$ into a network $N$ which computes the logical consequences of $T$. This provides also a learning system in the network that can be trained by examples using $T$ as background knowledge. In agents endowed with neural computation, induction is typically seen as the process of changing the weights of a network in ways that reflect the statistical properties of a dataset, allowing for generalisations over unseen examples. In the same setting, deduction is the neural computation of output values as a response to input vectors (encoding stimuli from the environment) given a particular set of weights. Such network computations have been shown equivalent to a range of temporal logic formalisms \cite{lamb_2007}. Based on this approach, an agent architecture called {\sc Neural Symbolic Cognitive Agent} ({\sc NSCA}) has been proposed in \cite{depenning_2011}. {\sc NSCA} uses temporal logic as theory $T$ and a Restricted Boltzmann Machine (RBM) as ANN $N$. An RBM is a partially connected ANN with two layers, a visible layer $V$ and a hidden layer $H$, and symmetric connections $W$ between these layers \cite{smolensky_1986}. 

An RBM defines a probability distribution $P(V=v,H=h)$ over pairs of vectors $v$ and $h$ encoded in these layers, where $v$ encodes the input data in binary or real values and $h$ encodes the posterior probability $P(H|v)$. Such a network can be used to infer or reconstruct complete data vectors based on incomplete or inconsistent input data and therefore implement an auto-associative memory or an autoencoder. It does so by combining the posterior probability distributions generated by each unit in the hidden layer with a conditional probability distribution for each unit in the visible layer. Each hidden unit constrains a different subset of the dimensions in the high-dimensional data presented at the visible layer and is therefore called an expert on some feature in the input data. Together, the hidden units can form a ``Product of Experts'' model that constrains all the dimensions in the input data. 

{\sc NSCA} is capable of (i) performing learning of complex temporal relations from uncertain observations, (ii) reasoning probabilistically about the knowledge that has been learned, and (iii) representing the agent's knowledge in a logic-based format for validation purposes. This is achieved by taking advantage of neural learning to perform robust learning and adaptation, and symbolic knowledge to represent qualitative reasoning. {\sc NSCA} was validated in a training simulator employed in real-world scenarios, illustrating the effective use of the approach. The results show that the agent model is able to learn to perform automated driver assessment from observation of real-time simulation data and assessments by driving instructors, and that this knowledge can be extracted in the form of temporal logic rules \cite{depenning_2011}.

\subsection{Mechanisms of Neural-Symbolic Integration Exemplified} Using a Recurrent Temporal Restricted Boltzmann Machine (RTRBM) \cite{sutskever_2009} as specialised variant of the general RBM concept, {\sc NSCA} encodes temporal rules in the form of hypotheses about beliefs and previously-applied rules. This is possible due to recurrent connections between hidden unit activations at time $t$ and the activations at time $t-1$ in the RTRBM. Based on the Bayesian inference mechanism of the RTRBM, each hidden unit $H_j$ represents a hypothesis about a specific rule $R_j$ that calculates the posterior probability that the rule implies a certain relation in the beliefs $b$ being observed in the visible layer $V$, given the previously applied rules $r^{t-1}$ (i.e. $P(R|B=b,R^{t-1}=r^{t-1})$). From these hypotheses the RTRBM selects the most applicable rules $r$ using random Gaussian sampling of the posterior probability distribution (i.e. $r  \propto P(R|B=b,R^{t-1}=r^{t-1})$) and calculates the conditional probability or likelihood of all beliefs given the selected rules are applied (i.e. $P(B|R=r)$). The difference between the observed and inferred beliefs can be used by {\sc NSCA} to train the RTRBM (i.e. update its weights) in order to improve the hypotheses about the observed data. For training, the RTRBM uses a combination of Contrastive Divergence and backpropagation through time. 

In the spirit of \cite{bratman_1999}'s Belief, Desire, Intention (BDI) agents, the observed data (e.g. simulation data or human assessments) are encoded as beliefs and the difference between the observed and inferred beliefs are the actual implications or intentions of the agent on its environment (e.g. adapting the assessment scores in the training simulator). The value of a belief represents either the probability of the occurrence of some event or state in the environment (e.g. $Raining=true$), or a real value (e.g. $Speed=31.5$). In other words, {\sc NSCA} deals with both binary and continuous data, for instance, by using a continuous stochastic visible layer \cite{chen_2003}. This improves the agent's ability to model asymmetric data, which in turn is very useful since measured data coming from a simulator is often asymmetric (e.g. training tasks typically take place in a restricted region of the simulated world). 

Due to the stochastic nature of the sigmoid activation functions, the beliefs can be regarded as fuzzy sets with a Gaussian membership function. This allows one to represent vague concepts like $fast$ and $slow$, as well as approximations of learned values, which is useful when reasoning with implicit and subjective knowledge \cite{sun_1994}.

The cognitive temporal logic described in \cite{lamb_2007} is used to represent domain knowledge in terms of beliefs and previously-applied rules. This logic contains several modal operators that extend classical modal logic with a notion of past and future. To express beliefs on continuous variables, this logic is extended with the use of equality and inequality formulae (e.g. $Speed<30$, $Raining=true$). As an example, consider a task where a trainee drives on an urban road and approaches an intersection. In this scenario the trainee has to apply a $yield-to-the-right$ rule. Using the extended temporal logic, one can describe rules about the conditions, scenario, and assessment related to this task. In rules (1) to (4) in Table~\ref{table1}, $\diamond A$ denotes ``$A$ is true sometime in the future'' and $A\boldsymbol{S}B$ denotes ``$A$ has been true since the occurrence of $B$''.

\begin{table}[h]
  \begin{compress}
    \bb{Conditions:}
    $(1)$ $(\p{Weather} > \p{good})$\newline
    $\p{meaning: the weather is at least good}$\newline
    \bb{Scenario:} 
    $(2)$ $\p{ApproachingIntersection} \wedge \diamond (\p{ApproachingTraffic} = \p{right})$\newline
    $\p{meaning: the car is approaching an intersection and sometime in the future traffic is approaching from the right}$\newline
    $(3)$ $((\p{Speed} > 0) \wedge \p{HeadingIntersection}) \boldsymbol{S} (\p{DistanceIntersection} < x) \rightarrow \p{ApproachingIntersection}$\newline
    $\p{meaning: if the car is moving and heading towards an intersection since it has been deemed close to the}\newline
    \p{intersection, then the car is approaching the intersection.}$\newline
    \bb{Assessment:}
    $(4)$ $\p{ApproachingIntersection} \wedge (\p{DistanceIntersection} = 0) \wedge (\p{ApproachingTraffic} = right) \wedge (\p{Speed} = 0) \rightarrow (\p{Evaluation} = good)$\newline
    $\p{meaning: if the car is approaching an intersection and arrives at the intersection when traffic is coming from the}\newline 
    \p{right and stops then the trainee gets a good evaluation}$\newline
      \end{compress}
      \vskip 0.1in
      \caption{Situations and assessments from a driving simulator scenario.\label{table1}}
\end{table}

Rule (4) is an example of an uncertain notion that is highly subjective (the distance $x$ at which a person is regarded as approaching an intersection is dependent on the situation and personal experience). When this rule is encoded in an RTRBM, it becomes possible to learn a more objective value for $x$ based on the observed behaviour of different people in various scenarios. This exemplifies a main objective of combining reasoning and learning.

The temporal logic rules are represented in {\sc NSCA} by setting the weights of the connections in the RTRBM. Therefore the rules need to be translated to a form that relates only to the immediately previous time step (denoted by the temporal operator $\bullet$). A transformation algorithm for this is also described in \cite{lamb_2007}. Then we can encode any rule as a stochastic relation between the hidden unit that represents the rule, the visible units that represent the beliefs, and the previous hidden unit activations that represent the applied rules in the previous time step. For example, the rule $\alpha \boldsymbol{S} \beta$ can be translated to the following rules: $\beta \rightarrow \alpha \boldsymbol{S} \beta$ and $\alpha \wedge \bullet(\alpha \boldsymbol{S} \beta) \rightarrow \alpha \boldsymbol{S} \beta$, where $\alpha$ and $\beta$ are modelled by visible units, $\alpha \boldsymbol{S} \beta$ by a hidden unit, and $\bullet(\alpha \boldsymbol{S} \beta)$ is modelled by a recurrent connection to the same hidden unit. \cite{pinkas_1995} shows how to map these logic-based propositions into the energy function of a symmetric network and how to deal with uncertainty by introducing a notion of confidence, called a penalty, as discussed further below.

\subsection{Applications of Neural-Symbolic Integration Exemplified} {\sc NSCA} has been developed as part of a three-year research project on assessment in driving simulators. It is implemented as part of a multi-agent platform for Virtual Instruction \cite{depenning_2008} and was used for an experiment on a driving simulator. In this experiment five students participated in a driving test consisting of five test scenarios each. For each attempt all data from the simulator (i.e. $43$ measurements, like relative positions and orientations of all traffic, speed, gear, and rpm of the student's car) and numerical assessments scores on several driving skills (i.e. vehicle control, economic driving, traffic flow, social-, and safe driving) that were provided by three driving instructors present during the attempts were observed by {\sc NSCA} in real-time. {\sc NSCA} was able to learn from these observations and infer assessment scores that are similar to those of the driving instructors \cite{depenning_2010,depenning_2011}. 

The {\sc NSCA} system has also been applied as part of a Visual Intelligence (VI) system, called CORTEX, in DARPA'€™s Mind'€™s Eye program \cite{donlon_2010}. This program pursues the capacity to learn generally-applicable and generative representations of actions between entities in a scene (e.g. persons, cars, objects) directly from visual inputs (i.e. pixels), and reason about the learned representations. A key distinction between Mind's Eye and the state-of-the-art in VI is that the latter has made large progress in recognising a wide range of entities and their properties (which might be thought of as the nouns in the description of a scene). Mind's Eye seeks to add the ability to describe and reason about the actions (which might be thought of as the verbs in the description of the scenes), thus enabling a more complete narrative of the visual experience. This objective is supported by {\sc NSCA} as it seeks to learn to describe an action in terms of entities and their properties, providing explanations for the reasoning behind it. Results have shown that the system is able to learn and represent the underlying semantics of the actions from observation and use this for several VI tasks, like recognition, description, anomaly detection, and gap-filling \cite{depenning_2012}. 

Most recently, NSCA has been included in an Intelligent Transport System to reduce CO2 emissions \cite{depenning_2014}. Results show that {\sc NSCA} is able to recognise various driving styles from real-time in-car sensor data and outperforms state-of-the-art in this area. In addition, {\sc NSCA} is able to describe these complex driving styles in terms of multi-level temporal logic-based rules for human interpretation and expert validation.

\subsection{{\sc NSCA} in a Nutshell}
In summary, {\sc NSCA} is an example for a cognitive model and agent architecture that offers an effective approach integrating symbolic reasoning and neural learning in a unified model. This approach allows the agent to learn rules about observed data in complex, real-world environments (e.g. expert behaviour for training and assessment in simulators). Learned behaviour can be extracted to update existing domain knowledge for validation, reporting, and feedback. Furthermore the approach allows domain knowledge to be encoded in the model and deals with uncertainty in real-world data. Results described in \cite{depenning_2010,depenning_2011} show that the agent is able to learn new hypotheses from observations and extract them into a temporal logic formula. But although results are promising, the model requires further evaluation by driving experts. This will allow further validation of the model in an operational setting with many scenarios, a large trainee population, and multiple assessments by driving instructors. Other on-going work includes research on using Deep Boltzmann Machines \cite{salakhutdinov_2009,tran_2012} to find higher-level rules and the application of an RTRBM to facilitate adaptive training. Overall, this work illustrates an application model for knowledge representation, learning, and reasoning which may indeed lead to realistic computational cognitive agent models, thus addressing the challenges put forward in \cite{valiant_2003,woolridge_2009}.

\section{Neural-Symbolic Integration in and for Cognitive Science: Building Mental Models} \label{mental_models}
Thus far, we elaborated on the general theoretical and conceptual foundations of neural-symbolic integration and discussed {\sc NSCA} as example of a successful implementation and application deployment of a neural-symbolic system. In what follows, we now return in more detail to some of the most pressing theoretical and applied core questions---including the binding problem, connectionist first-order logic (FOL) learning,  the combination of probabilities and logic in Markov Logic Networks, and the relationship between neural-symbolic reasoning and human-level AI---and describe our current state of knowledge, as well as potential future developments, in the following sections. We start with an introduction to the intimate relation between neural-symbolic integration and core topics from cognitive science.

To fully explain human cognition, we need to understand how brains construct and manipulate mental models. Clarification of this issue would inform the development of (artificially) intelligent systems and particularly neural-symbolic systems. The notion of mental model has a long history, perhaps most notably being used by Johnson-Laird to designate his theory of human reasoning \cite{johnson-laird_1983}. Taking partial inspiration from \cite{thagard_2010} and \cite{nersessian_2010}, we interpret a mental model as a cognitive representation of a real or imagined situation, the relationship amongst the situation's parts and, even perhaps, how those parts can act upon one another. These models will necessarily preserve the constraints inherent in what is represented, and their evaluation and manipulation has been argued to be key to human reasoning \cite{johnson-laird_1983}. For example, sight of an empty driveway and a scattering of broken glass, might conjure a mental representation (i.e. a model) of a felon breaking the window of your car, leading you to reason that your car had been stolen. As a more abstract example, when pondering on why the Exclusive Or problem is linearly inseparable in the context of network learning, one might visualise a two dimensional plane containing four regularly-placed crosses (two for true, two for false) and mentally consider placement of classification boundary lines. By exploring such a mental model, one might convince oneself that a line partitioning true and false outputs cannot be found.

It has particularly been argued that construction and inference on mental models might play a major role in abductive reasoning and, even, creativity \cite{thagard_2011}, our car theft example being a case of the former. Exactly how mental models and their manipulation are neurally realised, though, remains largely uncertain. Central to construction of a mental model is the formation of a combined (superordinate) whole representation from sets of (subordinate) part representations. For example, in the car theft example, a composite of your car and a hooded figure breaking a window might be created from memory and imagination of car and figure in isolation.

The most basic requirement of a neural realisation of such representation combination is a solution to the so called ``binding problem'', i.e. identification of a general neural mechanism to represent which individual assemblies of active neurones (e.g. those representing red Ford and those representing hooded figure) are associated and which are not (also see Section \ref{logical_neurons_binding}). Many proposals have been made for the brain's binding mechanism, e.g. temporal synchrony \cite{engel_2001}, conjunctive codes \cite{oreilly_2003,bowman_2007,rigotti_2010} and even convolution-based approaches \cite{thagard_2011}. Of these, conjunctive codes seem to involve the smallest step from traditional rate-coded ANNs, e.g. spiking and oscillatory dynamics are not assumed, and would thus seem most naturally integrable with neural-symbolic systems as currently formulated.

Under the conjunctive codes approach, it is assumed that units that respond selectively to the coactivation of multiple item representations are available. Such conjunctive units might reside in a widely accessible binding resource, such as, say, the Binding Pool in \cite{bowman_2007}. Two challenges to conjunctive code binding approaches (and, by extension, to neural representation of mental models) are, (i) scalability, and (ii) novel conjunctions:
\begin{itemize}
\item A binding resource that exhaustively enumerates all possible combinations of representations as unique (localist) units does not scale. Thus, realistic conjunctive codes typically assume that the binding resource employs distributed representations, which provide more compact, and indeed scalable, codings of binding associations, see for example, the explorations of this issue in \cite{rigotti_2010} and \cite{wyble_2006}.
\item We often experience representation combinations that we have never experienced before. A standard example is the proverbial blue banana or, in the context of the car theft example, it is likely that a mental model conjoining a hooded figure, a red Ford, and a particular driveway would never have been previously experienced. (Indeed, the fact that every experience must---necessarily---have a first occurrence makes the point.) As a result, conjunctive coding approaches propose randomly preconfigured binding resources with such discrimination capacity that they can effectively conjoin any combination of item representations that might arise in the future; the model of Frontal lobe function in \cite{rigotti_2010} is such an approach.
\end{itemize}
Neural conjunctive coding methods of this kind would, then, seem prerequisite to brain-based theories of mental models.

Clearly, a capacity to construct mental models is of limited value unless it helps us to make decisions and, in general, to reason, e.g. to determine that a crime has been committed and that the Police should be called. Two ways in which modern cognitive neuroscience suggests such reasoning could arise is through either statistical inference, or deliberative exploration. The former of these may employ statistical inference based upon pattern matching, in the manner of classic ANNs, as reflected in use of the descriptor ``associative learning'' in this context. In contrast, the latter would be more explicitly deliberative in nature and, accordingly, would involve the serial exploration of a space of hypothetical possibilities. Such a process might most naturally be framed in terms of classic symbolic computation and, perhaps, search in production system architectures \cite{newell_1994}. In addition, the statistical inference process might be viewed as more opaque to conscious awareness than deliberation.

Dual process theories of reasoning, e.g. \cite{evans_2003}, have argued that a subdivision of this kind is fundamental to human reasoning. With regard to our main interest here, though, the second of these forms of reasoning, deliberative exploration, seems most naturally to embrace the notion of mental model, with its connotation of forming conscious hypothetical representations.

Modern cognitive neuroscience also emphasises the role of emotion/body-state evaluations in reasoning. In particular, emotion may be viewed as providing an assessment of the evaluative quality of a situation or alternative, much in the manner of \cite{damasio_2008}'s somatic markers. But importantly, theories of affect, such as, say, appraisal theory \cite{scherer_1999}, suggest a multidimensional space of emotionally-charged feelings, richer than that required of simple reward and punishment or utility function. For instance, both disgust and fear are negatively valenced, marking, if you like, punishing experiences; however, the aversive responses to the two are typically quite different. In particular, only fear is likely to engage highly urgent flight or fight responses, associated with immediate physical risk. Indeed, it could be argued that a key function of emotions is to mark out response types and, at least in that sense, the distinctions between emotions on many dimensions are cognitively significant. This is relevant to cognitive and neuroscientific accounts of neural-symbolic integration in that mental models might, then, be viewed as ``stamped'' with affective evaluations, which would be bound (in the manner previously discussed) to nonaffective portions of a mental model, yielding a rich affect-coloured high dimensional representation. 

\subsection{Cognitive Neuroscience and Neural-Symbolic Methods}

Cognitive neuroscience (Gazzaniga, Ivry et al. 1998) is a major current scientific project, with the explanation of cognitive capacities (such as, perception, memory, attention, and language) in terms of their implementation in the brain as central objective. Computational modelling (especially when ANN-based) is a key ingredient of the research programme. Conceptually, such modelling provides the ``glue'' between cognition and the brain: it enables concrete (disproveable) explanations to be formulated of how cognitive capacities could be generated by the brain. In this sense, models that explain cognitive behaviour as well as neuroimaging data do a particularly good job of bridging between the cognitive and the neural. Examples of such models include (Bowman 2006; Chennu, Craston et al. 2009; Cowell and Cottrell 2009; Craston, Wyble et al. 2009), and the whole Dynamic Causal Modelling project (Friston, Harrison et al. 2003) can be seen to have this objective. 

This then brings to the fore the suitability of different cognitive modelling paradigms in Cognitive Neuroscience; that is, how should computational models in Cognitive Neuroscience be formulated? If we consider the cognitive and brain sciences in very broad definition, although computational models of many varieties have been employed, there are two dominant traditions. The first of these is symbolic modelling, which would most often be formulated in Production System architectures, such as {\sc SOAR} \cite{newell_1994}, {\sc EPIC} \cite{kieras_1999} and {\sc ACT-R} \cite{anderson_2014,anderson_2008}. In contrast, the second tradition is network modelling, ranging from abstract connectionist, e.g. \cite{rumelhart_1986}, to neurophysiologically detailed approaches, e.g. \cite{hasselmo_1997,fransen_2006}. From the earliest computation-based proposals for the study of mind and brain, which probably date to the 1950s, pre-eminence has oscillated between symbolic and neural network-based. However, modern interest in the brain and mapping cognition to it, has led to a sustained period of neural network pre-eminence and, certainly in the cognitive neuroscience realm, symbolic modelling is now rare; the most prominent exceptions being \cite{anderson_2008,taatgen_2009}.

There does though remain a minority who dissent from the neural networks emphasis. Some have taken a very strongly anti-neural networks perspective, e.g. \cite{fodor_1988}, arguing that thought is fundamentally symbolic---indeed linguistic, c.f. \cite{fodor_2001}---and that it should thus necessarily be studied at that level. Computer systems are often taken as a justification for this position. It is noted that software is the interesting level, which governs a computer's functional behaviour at any moment. In contrast, it is argued that the mapping to hardware implementation, e.g. compilation down to assembly code, is a fixed predefined transformation that does not need to be reprogrammed when the algorithm being run is changed. That is, the algorithm being executed is determined by software, not the mapping to hardware, or to paraphrase, in computer systems, software is where the functional complexity resides. The key point being that software is symbolic. If one views cognition as marked by its functional richness, one might then conclude that cognition is most appropriately studied symbolically and that the mapping to the brain is a fixed uninteresting implementation step, which does not reflect the richness and variety of cognitive experience and behaviour.

This position does, of course, hinge upon an analogy between the structure of computer systems and the structure of the mind-brain. Many would reject this analogy. Indeed, the majority of those that express dissent from network modelling, take a less extreme line. While acknowledging the relevance of brain implementation, they argue that ANNs are expressively limited as a cognitive modelling method, or---at least---emphasise that ANNs do not lead to natural modelling of certain cognitive functions. The following are cases in point.

\textbf{(i) Rule-guided problem solving:} Duncan has explored the human capacity to encode and apply sequences of rules governing how to perform mental tasks \cite{Duncan_2008}. Such \emph{task rules} state that fulfilment of a set of properties mandates a response of a particular kind. Furthermore, these properties and the relationships between them could be quite complex, e.g. ``if all the letters on one side of the screen are in capitals and an arrow points diagonally up to the right on the other side then respond in the direction indicated by the big triangle''.
Duncan and co-workers have shown that performance on their task rule experiments is strongly correlated with fluid intelligence and thus, with IQ. Performance is also associated with activation of brain areas believed to be involved in effortful task-governed behaviour \cite{dumontheil_2011}. Thus, these task rule experiments seem to be revealing the neurocognitive underpinnings of intelligence and problem solving; a key concern for science.
A natural way to think about these task rules is as symbolic operations with logic preconditions and actions that define a particular response. In this context, participants are seeking to correctly evaluate preconditions against presented stimuli and apply the corresponding (correct) action. Clearly, this perspective could be directly reflected in a production system model.
Finally, one of the key questions for task rule research is characterising how errors are made. To a large extent, errors seem to arise from preconditions and/or actions migrating between operations (i.e. between rules). Again, such errors could naturally be modelled in a production system formulation in which preconditions and actions could misbind to one another.

\textbf{(ii) Central Executive Function:} The notion of a centralised control system that guides thought (a Central Executive) is common in information processing theories of cognition, c.f. Baddeley and Hitch's working memory model \cite{baddeley_2000}, Shallice's Supervisory Attentional System \cite{Norman_1986} and the Central Engine in Barnard's Interacting Cognitive Subsystems \cite{barnard_1999}. The Central Executive is, for example, hypothesised to direct sensory, perceptual, and motor systems, particularly overriding prepotent (stereotyped) responses. Accordingly, it would be involved in mapping the organisms current goals to a strategy to realise those goals (a, so called, task set). In addition, it would impose that task set upon the brain's processing through centralised control. Indeed, the Central Executive would be involved in encoding and applying task rules of the kind just discussed, and also conscious exploration of sequences of alternatives in search of a strategy to obtain a goal, i.e. what in AI terms would be called planning. Neurophysiological localization of the Central Executive has focused on the Frontal lobe and particularly problem solving deficits arising from frontal lobe damage \cite{Norman_1986}. 
A number of the most prominent workers in this field have formulated their models in symbolic production systems, rather than ANNs. One such approach employs the {\sc COGENT} high-level modelling notation \cite{shallice_2011}, in which components, coded in production systems, execute in parallel, subject to intercomponent interaction. In particular, symbolic AI methods seem well suited to modelling some classic Central Executive (frontal lobe) tasks such as, Towers of Hanoi, Towers of London, and planning in general \cite{shallice_2011}.

\textbf{(iii) Syntactic structures:} A long running debate in psycholinguistics has contrasted rule-based and association-based explanations of language morphology, with Pinker typically carrying the flag for the former and McClelland and co-advocates for the latter \cite{pinker_2002}. The hottest debate has been focused on a particularly distinctive (U-shaped) profile of developmental data arising when children learn the English past tense inflection; i.e. when they determine that addition of an ``ed'' suffix enforces a past tense interpretation (e.g. ``stay'' to ``stayed''), subject to many exceptions (e.g. ``run'' to ``ran''). Although focused on one specific aspect of language, this debate serves as a sounding-board for a broader consideration: are the regular syntactic and morphological aspects of language in general, best viewed as symbolic rules or as emergent from the training of a Parallel Distributed Processing (PDP)-style network. This debate is specific to regular aspects of language; there is more agreement that exceptions could be naturally handled by PDP networks. 
The critical point for the connectionists is that, in the PDP case, there is no sense to which any particular rule or, indeed, an abstract notion of rule, is built into the network a priori; the network, in a sense, ``discovers'' regularity; indeed, if you like, the concept of regularity. Although the rules vs. associations debate will surely run and run, in terms of our focus here, it demonstrates an---at the least perceived---mismatch between classic connectionist approaches and the characteristics of cognitive behaviour. Furthermore, the proposed response to this mismatch is a symbolic modelling metaphor, which could naturally be realised in classical AI methods, such as, production systems, logic programming, etc.

\textbf{(iv) Compositionality:} One characterisation of the problem connectionism suffers is that it does not naturally reflect the representational compositionality inherent to many cognitive capacities \cite{fodor_1988}. For example, when parts are combined into wholes, those parts, at least to a large extent, carry their interpretation with them, e.g. ``John'' in ``John loves Jane'' has, ostensibly, the same meaning as it does in ``Jane loves John''. Although different classes of ANNs adhere to this characteristic to different extents, it certainly seems that classic PDP-style models do not automatically generate representational compositionality. At least when taken on face value, PDP models exhibit position-specific coding schemes; that is, the same item (e.g. ``John'') would have to be separately learned in position X and position Y, with no automatic interpretational carry-over from one position to the other. Consequently, the item may have a very different interpretation in the two positions and, indeed, what has been learned about ``John'' from its appearance at position X (where it might have been seen frequently) would not automatically inform the understanding of ``John'' at position Y (where it might have been seen rarely, if at all). This does not fit with subjective experience: the first time we see ``John'' in a particular grammatical position, having seen it elsewhere frequently, does not, it would seem, manifest as a complete unfamiliarity with the concept of ``John''. 
Although not always explicitly framed in terms of compositionality, the intermittent attacks on PDP approaches by proponents of localist neural coding schemes have a similar flavour \cite{page_2000}. Indeed, localists' criticism of the slot-coding used in PDP models of word reading \cite{bowers_2002} is very much a critique of position-specific coding.

This dissatisfaction with connectionism suggests a symbolic perspective, since symbolic models are foundationally compositional in the fashion ANNs are argued not to be.
If one accepts this weight of argument, one is left with the perspective that while the link to the brain is certainly critical (indeed, the brain does only have neurones, synapses, etc., to process with), it may not be practical to map directly to neural implementation. Rather, cognitive modelling at an intermediate symbolic level may be more feasible.

In particular, ANNs have certainly made a profound contribution to understanding cognition. This though has been most marked for a specific set of cognitive faculties. For example, compelling models of sensory processes have been developed, e.g. stages of the ``what'' visual processing pathway (i.e. the ventral stream), c.f. \cite{li_2001,raizada_2003,serre_2007}; and face-specific processing within the what pathway, cf. \cite{burton_1990,cowell_2009}. Mature neural models of visual attention through both space and time have also been proposed, e.g. \cite{itti_1998,heinke_2003,bundesen_2005,bowman_2007}. There are also sophisticated neural models of memory systems and learning in general, e.g. \cite{bogacz_2003,Norman_2003,davelaar_2004}. Furthermore, many of these models are compelling in their neurophysiological detail.

However, as justified above, there are areas where neural modelling has had less success. In general terms, architectural-level neural models are scarce, i.e. models that are broad-scope, general-purpose, and claim relevance beyond a specific cognitive phenomenon. There are, though, many symbolic architecture-level models of mind, e.g. {\sc SOAR} \cite{newell_1994}---possible reasons for this situation have been discussed, among others, in \cite{bowman_2011}. It may also be the case that neural explanations become less compelling as cognitive functions become more high-level; that is, they are well suited to characterising ``peripheral'' systems, such as, vision, audition, and motor action, but they do less well as one steps up the processing hierarchy away from input and output. In particular, what Sloman would call deliberative processing \cite{sloman_1999}, which might embrace Duncan's task rule experiments, planning and, at least elements of, Central Executive function, are less naturally modelled with ANNs. To emphasise again, we are not saying that such cognitive capacities cannot in some fundamental sense be neurally modelled, but rather we are suggesting it might not be the most natural method of description.

As already suggested, a possible response to this situation is to consider whether mapping to neural substrate could be expedited by breaking the modelling problem into two steps: the first a symbolic modelling of cognitive behaviour and the second a mapping from symbolic to neural. From a philosophical perspective, one would be arguing that the symbolic intermediary more directly reflects the functional characteristics of the cognitive capacity in question, leading the modeller to a, somehow, more faithful model at the cognitive level. Furthermore, the suggestion would be that such a functional characterisation would also strongly constrain the neural implementation and vice-versa, via a direct mapping to ANNs, i.e. effectively a compilation step (which may nevertheless be necessary computationally for the sake of efficiency). 

The first, cognitive behaviour to symbolic model, step might very naturally yield some form of computational logic model, perhaps formulated in a production systems architecture, such as {\sc SOAR} \cite{newell_1994}, {\sc ACT-R} \cite{anderson_2014} or {\sc EPIC} \cite{kieras_1999}, or as a logic program. Either way, the second mapping would involve some form of computational logic to ANN transformation---exactly the topic of the neural-symbolic project. Thus, in this respect, application of the results of neural-symbolic research could make a significant contribution to Cognitive Neuroscience.

However, such an application raises a number of research topics. Perhaps the most significant of these is to develop mappings from computational logic to neurophysiologically plausible neural networks. This requires consideration of more complex activation equations that directly model ion channels and membrane potential dynamics (e.g. based upon Hodgkin-Huxley equations \cite{Hodgkin_1952}); it also suggests approaches that treat excitation and inhibition distinctly and employ more biologically-realistic learning, such as contrastive Hebbian learning, as arises in, say, O'Reilly and Munakata's Generalised Recirculation Algorithm. A good source for such neural theories is \cite{oreilly_2000}.

\section{Putting the Machinery to Work: Binding and First-Order Inference in a Neural-Symbolic Framework} \label{logical_neurons}
While it should have become clear that it seems plausible to assume that human cognition produces and processes complex combinatorial structures using neural networks, on the computer science and systems engineering side such systematic and dynamic creation of these structures presents challenges to ANNs and for theories of neuro-cognition in general \cite{fodor_1988,marcus_2003,vandervelde_2006,feldman_2013}. We now will zoom in on concrete work relating to the problem of representing such complex combinatorial structures in ANNs---with emphasis on FOL representations---while efficiently computationally learning and processing them in order to perform high-level cognitive tasks. Similar to the presentation of {\sc NSCA} in Section \ref{learning_and_reasoning} as application example for the theoretical and conceptual considerations from Section \ref{prolegomena}, the following shall---besides presenting a relevant body of work from neural-symbolic integration---provide a practical counterpart and grounding for some of the cognition and cognitive neuroscience-centred discussion of the previous Section~\ref{mental_models}.

Specifically, this section describes how predicate logic can efficiently be represented and processed in a type of recurrent network with symmetric weights. The approach is based on a (variable) binding mechanism which encodes compactly first-order logic expressions in activation. Processing is then done by encoding symbolic constraints in the weights of the network, thus forcing a solution to an inference problem to emerge in activation.

\subsection{A Computational View on Fodor, Pylyshyn, and the Challenge(s) for Modelling Cognition}

One of the computationally most influential and relevant positions relating cognition and neural-symbolic computation has been developed by \cite{fodor_1988}. There, two characteristics are deemed essential to any paradigm that aims at modelling cognition: (i) combinatorial syntax and semantics for mental representations and (ii) structure sensitivity of processes. The first characteristic allows for representations to be recursively built from atomic ones in a way that the semantics of a non-atomic structure is a function of the semantics of its syntactic components. The second characteristic refers to the ability to base the application of operations on a structured representation, on its syntactic structure. 

Several works partially answered their call, yet none could capture the full expressive power of FOL: Even with limited expressiveness, most attempts to address Fodor's and Pylyshyn's challenges are extremely localist, have no robustness to neural damage, have limited learning ability, have large size-complexity, and require ad hoc network engineering rather than using a general purpose mechanism which would be independent of the respective knowledge base (KB) \cite{anandan_1989,ballard_1986,garcez_2009,hoelldobler_1990,hoelldobler_1992,stolcke_1989,vandervelde_2006}. Additionally, only few systems used distributed representations---and even those which do typically suffer from information loss, and little processing abilities \cite{plate_1995,pollack_1990,stewart_2008}. 

\subsection{The Binding Problem Computationally Revisited}\label{logical_neurons_binding}

Despite the large body of work (non-exhaustively) summarised in Section \ref{mental_models} and the corresponding many attempts to approach Fodor's and Pylyshyn's criticism, computational neural modelling of high-level cognitive tasks and especially language processing is still considered a hard challenge. Many of the problems in modelling are taken to very likely be related to the already previously described ``binding problem''. Following the definition by \cite{feldman_2013}, in short the general binding problem concerns how items, which are encoded in distinct circuits of a massively parallel computing device, can be combined in complex ways, for various cognitive tasks. In particular, binding becomes extremely challenging in language processing and in abstract reasoning \cite{barrett_2008,jackendorff_2002,marcus_2003,vandervelde_2006}, where one must deal with a more demanding, special case: variable binding. The underlying question is narrowed down to: How are simple constituents glued to variables and then used by several and distinct parts of a massively parallel system, in different roles? Following \cite{jackendorff_2002}, the binding problem can be further decomposed into four constitutive challenges: (i) the massiveness of the binding problem, (ii) the problem of multiple instances of a fact, (iii) the problem of associating values to variables and (iv) the relation between bindings in short-term working memory (WM) as opposed to bindings in long-term memory (LTM). When jointly looking at all four, the binding mechanism seems to be the fundamental obstacle as each individual item depends on it. 
But even if a binding mechanism should exist and enables the representation of complex structures in neurone-like units, it is not clear how they are processed in order to achieve a goal and how the procedural knowledge needed for such processing is encapsulated in the synapses. Specifically, in the computational case, when FOL expressions shall be processed using just neural units with binary values, how should the respective formulae be encoded? How are neural-ensembles, representing simple constituents, glued together to form complex structures? And how are these structures manipulated for unification and ultimately for reasoning? 

As a first step, a binding mechanism is required that will allow to ``glue'' together items in a KB, such as predicates, functions, constants, and variables, and then apply it in manipulating these structures. Correspondingly, several attempts have been made to tackle the variable binding problem \cite{anandan_1989,barrett_2008,browne_1999,shastri_1993,vandervelde_2006}. But while the respective approaches have greatly contributed to our understanding of possible mechanisms at work, it turned out that they each still have signifiant shortcomings related to limited expressiveness, high space requirements, or central control demands.

\subsection{Inference Specifications as Fixed Points of ANNs}
 
Instead of developing a bottom-up, modular solution to performing reasoning with ANNs, let us examine the opposite approach: specifying what should be the possible results of a reasoning process as fixed points (or stable states) of an ANN. In every ANN that reaches equilibrium, converging onto a stable state (instead of oscillating), the fixed points of the network---given clamped inputs---can be associated with solutions for the problem at hand. In particular, ANNs with a symmetric matrix of weights such as, for instance, the already discussed Boltzmann machines from Section \ref{learning_and_reasoning}, perform gradient descent in an energy function, whose global minima can be associated with solutions to the problem. Although symmetric (energy minimisation) networks are the straightforward architecture choice, other (non-symmetric) architectures may be used as long as stable states emerge and those stable states correspond to solutions of the problem at hand. When using symmetric networks, the general strategy is to associate the problem-solutions with the global minima of the energy equation that corresponds to the network. In particular, as the ``problem'' is FOL inference, valid inference chains are the solutions, which are mapped into the global minima of the energy function.  

It has been shown that high-order symmetric ANNs (with multiplicative synapses) are equivalent to ``standard'' symmetric ANNs (with pairwise synapses) with hidden units \cite{pinkas_1991}. In the corresponding simulations, high-order variants of Boltzmann Machines are therefore used, as they are faster to simulate and have a smaller search space than the ``standard'' Boltzmann machines with hidden units and only pair-wise connections. The declarative specifications (constraints) characterising valid solutions are compiled (or learned) into the weights of a higher-order Boltzmann machine. The first-order KB is either compiled into weights, or it is clamped onto the activation of some of the visible neurones. A query, short-term facts, and/or optional cues may also be clamped onto the WM. Once the KB is in its place (stored as long-term weights) and queries or cues are clamped onto the WM, the ANN starts performing its gradient descent algorithm, searching for a solution for the inference problem that satisfies the constraints stored as weights. When the network settles onto a global minimum, such a stable state is interpreted as a FOL inference chain proving queries using resolution-based inference steps \cite{pinkas_nips_1991,lima_thesis_2000,pinkas_2012,pinkas_2013}.

\subsubsection{ANN Specifications Using Weighted Boolean Constraints}
 
The first step in the process is the specification of the problem (e.g., FOL inference) as a set of weighted propositional logic formulae. These formulae represent Boolean constraints that enforce the neurones to obtain zero/one assignments that correspond exactly to the valid FOL inference solutions. 

\begin{figure}
\begin{center}
\includegraphics[width = 0.5\textwidth]{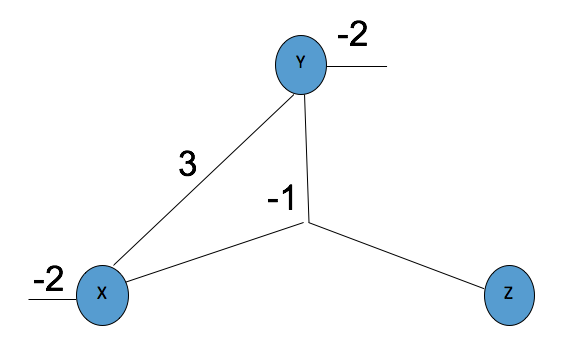}
\caption{A symmetric 3-order network, characterised by a 3-order energy function: $XYZ-3XY+2X+2Y$. Minima of this energy function are fixed points of the network. This network searches for satisfying solutions for the weighted conjunctive normal form (CNF): $(\neg X \vee \neg Y \vee Z) \wedge (X \vee Y)$. Note that the clauses of the weighted CNF are augmented by penalties reflecting the importance of each constraint.}\label{symmetric_network}
\end{center}
\end{figure}

Satisfying a propositional formula is the process of assigning truth-values (true/false) to the atomic variables of the formula, in such a way that the formula turns out to be true. Maximally satisfying a weighted set of such formulae is finding truth-values that minimise the sum of the weights of the violated formulae. By associating the truth-value true with 1 and false with 0, a quadratic energy function is created, which basically calculates the weighted sum of the violated constraints. Conjunctions are expressed as multiplicative terms, disjunctions as cardinality of union, and negations are expressed as subtraction from 1. In order to hint the network about how close it is to a solution, the conjunction of weighted formulae is treated as a sum of the weights of Boolean constraints that are not satisfied. It is possible to associate the problem of maximal-weighted-satisfiability of the original set of weighted formulae with finding a global minimum for the energy equation that corresponds to the problem specification \cite{pinkas_1991}. In other words, not satisfying a clause would incur an increase of the system's energy. One interesting and powerful side effect of such a mapping is the possibility of partially satisfying an otherwise unsatisfiable set of formulae. This phenomenon is used for specifying optimal and preferred solutions as in: finding a shortest explanation or proof (parsimony), finding most general unifications, finding the most probable inference or most likely explanation, etc. When different levels of weights are associated with the formulae of the KB, a non-monotonic inference engine is obtained based on Penalty Logic. In fact, in \cite{pinkas_1995} equivalence has been shown between minimising the energy of symmetric ANNs and satisfiability of penalty logic, i.e., every penalty logic formula can be efficiently translated into a symmetric ANN and vice versa. Figure \ref{symmetric_network} gives an example for a network instantiation of a satisfiability problem.

Based on these results, a compiler has been built that accepts solution specifications as a form of index-quantified penalty logic formulae and generates ANNs that seek for maximally satisfying solutions for the specified formulae. This compiler can then be used to generate ANNs for a variety of FOL retrieval, unification, and inference problems, involving thousands of neural units and tens of thousands of synapses. Alternatively to compilation, these specification constraints could also be learned (unsupervised) using an anti-Hebbian learning rule, whenever a fixed point is found that does not satisfy some constraints \cite{pinkas_1995}.  

\subsubsection{ANN Specifications of FOL Inference Chains}

There are two possible approaches to logical reasoning, one that uses Model Theory and the other Proof Theory. Informally, Model Theory confirms that a statement could be inferred by checking all possible models that satisfy a KB, while Proof Theory applies inference steps to generate a proof for the statement in question. Most attempts at doing FOL inferences in ANNs---or using propositional satisfiability in general---employ model-theoretic techniques based on grounded KBs \cite{clarke_2001,garcez_2002,domingos_2008,hoelldobler_1992}. However, grounding results in either exponential explosion of the number of boolean variables (neurones) or severe limitations to the expressive power of the FOL language (e.g. no function symbols and no existential quantifiers). Proof Theory techniques may use ungrounded FOL formulae and are more efficient in size. Another consequence of having a proof for a query statement is that this proof can be seen as an explanation for that statement, which is not always possible to obtain when model-theoretic methods are used. A proof-theoretic approach was, for instance, used by \cite{pinkas_nips_1991} and allows for a variety of inference rules (e.g. Resolution, Modus Ponens, etc.) with either clamped KB or with the KB stored in weights. In \cite{lima_thesis_2000}, a FOL theorem prover has been studied with resolution and a knowledge base that is clamped entirely in activation. In \cite{pinkas_2013} most general unification with occurrence checking is implemented using a more recent  binding mechanism with a reduced size complexity.

After the query and/or cues had been clamped onto the WM, the network is set to run until a solution emerges as a global minimum of the network's energy function. This stable state may be interpreted as a proof or a chain of FOL clauses. Each clause is obtained by either retrieving it from the LTM, by copying it from previous steps (optionally with collapse), or by resolving two previous clauses in the chain.  In every step, the need to unify resolved literals may cause literals to become more specific by variable binding. It is possible to construct an ANN that searches for proofs by refutation (a complete strategy that derives a contradiction when the negated query is added to the KB) or directly infer a statement. Directly inferring the conclusion, though not a complete strategy, bears more resemblance to human reasoning, and seems more cognitively plausible.

Variations of the above basic inference mechanism use competition among proofs and can be used for inferring the most likely explanation for a query, most probable explanation, most probable inference, least expensive plan or most preferable action (as in Asimov's robotic laws).

\subsection{A Fault Tolerant Mechanism for Dynamic Binding}

A compact binding mechanism is at the foundation of the described FOL encoding and of the general purpose inference engine. The binding technique is the basic mechanism upon which the WM is constructed and it is this mechanism that enables the retrieval of the needed knowledge from LTM and then unification and inference. 

Contrary to temporal binding \cite{shastri_1993}, it uses spatial (conjunctive) binding which captures compositionality in the stable state of the neurones and is not sensitive to timing synchronisation or time slot allocation. Although several conjunctive (non-temporal) binding mechanisms have been proposed, the one in \cite{pinkas_2012} compactly uses only few ensembles of neurones, each of which can bind any of the many knowledge items stored in the LTM. This property has three side effects: (i) reduction in the number of neurones (units), as binders are only needed to represent a proof (and not to capture all the KB), (ii) fault tolerance, as each binding ensemble is general purpose and may take the place of any other failed binding ensemble, and finally (iii) high expressive power, as any labelled directed graph structure may be represented by this mechanism (FOL as a special case). 

This binding technique applies crossbars of neurones to bind FOL objects (clauses, literals, predicates, functions, and constants) using special neural ensembles called General Purpose (GP) binders. These GP-binders act as indirect pointers, i.e., two objects are bound if and only if a GP-binder points to both of them. Since GP-binders can point to other GP-binders, a nested hierarchy of objects may be represented using such binders pointing at the same time to objects/concepts and to other binders. In fact, arbitrary digraphs may be represented and, as special cases, nested FOL terms, literals, clauses, and even complete resolution-based proofs \cite{pinkas_nips_1991,lima_thesis_2000}. Manipulation of such compound representations is done using weighted Boolean constraints (compiled or learned into the synapses), which cause GP-binders to be dynamically allocated to FOL objects, retrieved from the LTM.  Since the GP-binders can be used to bind any two or more objects, the number of binders that is actually needed is proportional to the size of the proof. As a result only few such binders are actually needed for cognitively-plausible inference chains, causing the size complexity of the WM to be drastically reduced to $O(n\cdot log(k))$ where $n$ is the proof length and $k$ is the size of the KB \cite{pinkas_2013}. Interestingly, the fault tolerance of this FOL WM is the result of the generic nature of the binders, which is due to redundancy in synapses rather than to redundancy in units. GP-binders happen to address most of the challenges stated by \cite{jackendorff_2002} and summarised above.

\subsection{Examples of Inferences, Cues, and Deductive Associations}

To illustrate the GP-binder approach, assume a KB consisting of the following rules:

\begin{itemize}
\item For every $u$, there exists a mother of $u$ who is also a parent of $u$: $Parent(mother(u),u)$
\item For all $x,y,z$: $Parent(x,y) \wedge Parent(y,z) \rightarrow GrandParent (x,z)$
\item More KB items: facts and/or rules optionally augmented by weights
\end{itemize}

A general purpose resolution-based inference engine and a WM implemented in a variation of Boltzmann Machines is compiled (or learned) with the above KB stored in its synaptic connections.  When clamping the query: $GrandParent(V,Anne)$ onto some of the visible units in the WM,  while settling onto a global energy minimum, the network  is capable of binding the variable $V$ to the term representing ``the mother of the mother of Anne'' and finding the complete inference chain as a proof for that query:

\begin{enumerate}
\item Retrieving and unifying rule 1: $Parent(mother(Anne),Anne)$
\item Retrieving and unifying rule 2: $\neg Parent(mother(mother(Anne)),mother(Anne)) \vee$ $\neg Parent(mother(Anne),Anne) \vee$ $GrandParent(mother(mother(Anne)),Anne)$
\item Resolving 1 and 2: $\neg Parent(mother(mother(Anne)), mother(Anne)) \vee$\\$GrandParent(mother(mother(Anne)),Anne)$
\item Retrieving rule 1 again but with different unification (multiple instances):\\$Parent(mother(mother(Anne)), mother(Anne))$
\item QED by resolving 3 and 4: $GrandParent(mother(mother(Anne)),Anne)$
\end{enumerate}

FOL clauses, literals, atoms, and terms are represented as directed acyclic graphs while GP-binders act as nodes and labeled arcs. The GP-binders are dynamically allocated without central control and only the needed rules are retrieved from the KB, in order to infer the query (the goal). Other rules that are stored in the KB (in the LTM) are not retrieved into the WM because they are not needed for the inference.

The query may be considered as a kind of cue in the WM, that directs the network's ``attention'' to searching for the grandparent of Anne; yet, other cues may be provided instead or in addition to the query: for example, we can accelerate the system's convergence to the correct solution by specifying in the WM that rule 1 and/or rule 2 should be used, or, even better, by clamping these rules onto the WM activation prior to the network operation. 

This ANN inference mechanism could also be interpreted as an intelligent associative memory because cues generate a chain of deduction steps that leads to complex associations. For example, instead of a query, if just ``Anne'' and ``mother'' are clamped as cues in the WM this amounts to asking the system to infer something about the person ``Anne'' and the concept ``mother''. In this case the system will converge to one of several possible inferences (each with a known probability). For instance, the system may find that Anne has a parent who is her mother. Alternatively, it may find that Anne has a grandparent (the mother of her mother), or that Anne's mother has a mother and that mother also has a mother.  The likelihood of the possible emerging associations is determined by such factors as the weights of the LTM (i.e. penalties of the FOL KB items, stored in the synapses) and the depth of the inference chains.

\subsection{Can Neurones Be Logical? Fundamental Negative and Positive Results}

The just described work provides evidence that unrestricted (but memory-bound) FOL inference can be represented and processed by relatively compact, attractor-based ANNs such as the previously introduced Boltzmann machines. Nevertheless, in \cite{pinkas_dechter_1995} it has also been shown that distributed neural mechanisms---among which also Boltzmann Machines count---may have pathological oscillations in ANNs that have cycles and may never find a global solution. The related positive result is that when the ANN has no cycles, it is possible to have a neural activation function that guarantees finding a global solution. Once a cycle is added to the network, no activation function is able to eliminate all the pathological scenarios above. Unfortunately, the outlined FOL networks have many cycles and are therefore doomed to be incomplete even with a bound on the proof length. Luckily with some plausible assumptions on the daemon, one can guarantee that such pathological scenarios will not last forever. Another major obstacle for achieving good performance is that Boltzmann Machines and other symmetric networks use stochastic local search algorithms to search for a global minimum of their energy function.  Theoretically and practically, local minima may plague the search space, so such ANNs are not guaranteed to find a solution in a finite time. Current research is trying to solve this problem by using many variations of Hebbian learning methods and variations of Boltzmann machines to change the synapses of an ANN in a way that local minima are eliminated and a global minimum is found faster. As it turns out, the recent success of deep learning mostly also emerges from the same variations but from a more practical perspective. Learning styles of this type are called ``speed-up practicing'' as they aim to use learning as a strategy for self-improvement.

\section{Connectionist First-Order Logic Learning Using the Core Method} \label{connectionist_first-order_logic}
As has already become clear in the previous sections, one of the central problems at the core of neural-symbolic integration is the representation and learning of first-order rules (i.e. relational knowledge) within a connectionist setting. But there is a second motivation besides the incentives coming out of cognitive science and cognitive neuroscience and aiming at the modelling and (re-)creation of high-level cognitive processes in a plausible neural setting: The aim to create a new, neurally-inspired computational platform for symbolic computation which would, among others, be massively distributed with no central control, robust to unit failures, and self-improving with time.

In this and the following section we take a more technical perspective---to a certain extent contrasting the strongly cognitively-motivated Sections \ref{mental_models} and \ref{logical_neurons}---and focus on two prototypical examples for work on the computer science side of neural-symbolic integration, namely (i) the approximation of the immediate consequence operator $T_P$ associated to a FOL program $P$ in a neural learning context, and (ii) Markov Logic as probabilistic extension of FOL combining logic and graphical models. While relegating elaborations on the latter topic to the following Section \ref{markov_logic}, we start with the question of the consequence operator. 

Assume a FOL program containing the fact $p(X)$. Applying the associated $T_P$-operator once (to an arbitrary initial interpretation), leads to a result containing infinitely many atoms, namely all $p(X)$-atoms for every $X$ within the underlying Herbrand universe $U_L$. Thus, one has to approximate the operator because even a single application can lead to infinite results. And while in this simple example one might still be able to define a finite representation, the problem might become arbitrary complex for other programs. So-called rational models have been developed to tackle this problem \cite{bornscheuer_1996}. 

Unfortunately, there is no way to compute an upper bound on the size of the resulting rational representation. Because we are not aware of any other finite representation, we will in the following concentrate on the standard representation using Herbrand interpretations. In principle, there are two options to approximate a given operator. On the one hand, one can develop an approximating connectionist representation for a given accuracy. This approximation in space leads to an increasing number of hidden layer units in the resulting networks \cite{bader_2009,bader_2007}. Alternatively, a connectionist system can be constructed that approximates a single application of $T_P$ the better the longer it runs. This approximation in time was used in \cite{bader_2004} and \cite{bader_2005}. Networks constructed along the approximation in space approach are more or less standard architectures with many hidden layer units, while the others are usually very small but use non-standard architectures and units. 

\subsection{Feasibility of the First-Order Core Method and the Embedding of First-Order Rules into Neural Networks}

It is well known that multilayer feed-forward networks are universal approximators for all continuous functions on compact subsets \cite{funahashi_1989}. If a suitable way to represent first-order interpretations by (finite vectors of) real numbers could be found, then feed-forward networks may be used to approximate the meaning function of such programs. It is necessary that such a representation is compatible with both, the logic programming and the ANN paradigm. For this, one defines a homeomorphic embedding from the space of interpretations into a compact subset of the real numbers. \cite{hoelldobler_1999} showed the existence of a corresponding metric homeomorphism, and \cite{hitzler_2004} subsequently expanded this result to the characterisation of a topological isomorphism. Following these insights, one can use level mappings to realise this embedding. For some Herbrand interpretation $I$, a bijective level mapping $|\cdot |:B_L \rightarrow \mathbb{N}^+$, and $b>2$, the embedding function $\eta:B_L\rightarrow \mathbb{R}$ and its extension $\eta:I_L\rightarrow \mathbb{R}$ are defined as follows:

$$ \eta : B_L \rightarrow \mathbb{R} : A \mapsto b^{-|A|} \hspace{1cm} \eta : I_L \rightarrow \mathbb{R}: I \mapsto \sum_{A \in I} \eta(A)$$

Here, $C_b:=\{\eta(I) | I \in L\} \subset R$ is used to denote the set of all embedded interpretations. This results in a ``binary'' representation of a given interpretation $I$ in the number system with base $b$. After embedding an interpretation and looking at its representation within this number system one finds a $1$ at each position $|A|$ for all $A\in I$. As mentioned above, the embedding is required to be homeomorphic (i.e. to be continuous, bijective and to have a continuous inverse), because being homeomorphic ensures that $\eta$ is at least a promising candidate to bridge the gap between logic programs and connectionist networks. One can construct a real valued version $f_P$ of the immediate consequence operator $T_P$ as follows:
$$f_P : C_b \rightarrow C_b : X \mapsto \eta(T_P(\eta^{-1}(X)))$$
Because $\eta$ is a homeomorphism, $f_P$ is a correct embedding of $T_P$ in the sense that all structural information is preserved. Furthermore, $\eta$ is known to be continuous which allows to conclude that $f_P$ is continuous for a continuous $T_P$. Based on these insights, one can construct connectionist networks to approximate $T_P$. As mentioned above there are in principle two options for the approximation---in space and in time. The embedding $\eta$ introduced above allows for both, as discussed in detail in \cite{bader_nesy_2005,bader_2007,bader_2004,bader_2005}. In \cite{bader_thesis_2009}, methods have been described in full detail to construct standard sigmoidal networks, as well as radial basis function networks for any given accuracy of approximation. That is, if an interpretation $I$ is fed into the network, the resulting output can be interpreted as an interpretation $J$ and the distance between $J$ and $T_P (I)$ is limited by a given $\epsilon$.

So far we have been concerned with the approximation of a single application of the immediate consequence operator only. More interesting is whether the repeated application of the approximating function converges to a fixed point relating to the model of the underlying program. Convergence can be shown by showing that a function is contractive and by employing the Banach contraction mapping principle. Unfortunately, it turns out that the approximating functions described above are not contractive on $R$ in general, but they are contractive if restricted to $C_b$. Therefore, one can conclude that not only the meaning function $T_P$ can be approximated using a connectionist network, but also the iteration of the approximation converges to a state corresponding to a model of the underlying program provided that $T_P$ itself is contractive.

\subsection{Learning in a First-Order Setting}

Within a propositional setting, learning has been shown to be possible \cite{garcez_2002} and to be of advantage \cite{bader_2008}. As in the propositional case, training data is used to adapt the weights and structure of randomly initialised networks. An approach built upon vector-based networks has first been proposed in \cite{bader_2007} along with a specifically tailored learning method. To show the applicability of that approach, a network had randomly been initialised and pairs of interpretations and corresponding consequence (i.e. pairs of $I$ and $T_P(I)$), had been used as training samples. After convergence of the training process the network was evaluated by feeding random interpretations into the network and iterating the computation by feeding the outputs back into the input layer. The resulting sequence of interpretations was found to converge to the model of the underlying program $P$. Hence it can be concluded that the network indeed managed to learn the immediate consequence operator sufficiently accurate.

From a more practical perspective (acknowledging the above difficulties, yet recognising their scientific, representational value), some have opted for connecting neural computation with the work done in Inductive Logic Programming \cite{muggleton_1994,basilio_2001}. Initial results, that take advantage of ILP methods for efficient FOL learning called propositionalisation \cite{krogel_2003}, have only started to emerge, but with promising results (cf., among others, the Connectionist ILP system by \cite{franca_2014}, or the work by \cite{sourek_2015,deraedt_2016}). Also within a more practical approach, an extension of the above fixed-point method to multiple networks, as shown in Figure~\ref{neural-symbolic_system}, has been proposed in \cite{garcez_2009}. Such an extension, called connectionist modal logic, is of interest here in relation to FOL learning because it has proven capable of representing and learning propositional modal logics, which in turn are equivalent to the fragment of FOL with two variables. Propositional modal logic has been shown to be robustly decidable \cite{vardi_1996} and it can be considered, as a result, a candidate for offering an adequate middle ground between the requirements of representation capacity and learning efficiency.

\section{Markov Logic Networks Combining Probabilities and First-Order Logic} \label{markov_logic}
As a second prototypical example for the more technical side of neural-symbolic integration besides the connectionist first-order learning application from the previous section, we now turn our attention to Markov logic networks. Markov logic is a probabilistic extension of FOL \cite{richardson_2006,domingos_2009}. Classical logic can compactly model high-order dependencies. However, at the same time it is deterministic and therefore brittle in the presence of noise and uncertainty. The key idea of Markov logic is to overcome this brittleness by combining logic with graphical models. To that end, logical formulae are softened with weights, and effectively serve as feature templates for a Markov network or Markov Random Field (MRF)~\cite{taskar_2004}, i.e. an undirected graphical model that allows arbitrary high-order potentials, similar to (but potentially much more compact than) a high-order Boltzmann distribution.

Concretely, a Markov logic network (MLN) is a set of weighted FOL formulae. Together with a set of constants, it defines a Markov network with one node per ground atom (a predicate with all variables replaced by a constant) and one feature per ground formula. The weight of a feature is the weight of the first-order formula that originated it. The probability of a state $x$ in such a network is given by $P(x) = \frac{1}{Z} \exp(\sum_i w_i f_i(x))$, where $Z$ is a normalisation constant, $w_i$ is the weight of the $i$-th formula, $f_i = 1$ if the $i$-th formula is true, and $f_i = 0$ otherwise.

\begin{figure}
\begin{center}
\includegraphics[width = 0.65\textwidth]{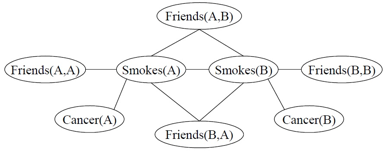}
\caption{A Markov network with two constants Anna ($A$) and Bob ($B$).}\label{markov_network}
\end{center}
\end{figure}

As an example, Table~\ref{table2} shows an MLN with two weighted formulae, and Figure~\ref{markov_network} shows the corresponding Markov network with two constants, Anna ($A$) and Bob ($B$).

\begin{table}
\begin{tabular}{|l|l|l|}
\hline
\textbf{English} & \textbf{First-Order Logic} & \textbf{Weight}\\
\hline
Smoking causes cancer & $\forall x: Sm(x) \rightarrow Ca(x)$ & $1.5$\\ 
\hline
Friends share smoking habits & $\forall x \forall y: Fr(x,y) \wedge Sm(x) \rightarrow Sm(y)$ & $1.1$\\
\hline
\end{tabular}
\caption{\label{table2}}
\end{table}

% NEW DBL TODO: Add a few citations.  Split into 2+ paragraphs.
% Revision pass.  Double-check inference stuff?
Exact and approximate inference methods for MRFs and other graphical
models can also be applied to MLNs.  Popular choices include variable
elimination or junction trees for exact inference and message-passing
or Markov chain Monte Carlo methods for approximate inference.
However, compared to standard MRFs, MLNs introduce new challenges and
opportunities for efficient reasoning.  One challenge is that even
modestly sized domains can yield exceptionally large and challenging
ground Markov networks.  For example, in a social network domain such
as Table~\ref{table2}, the number of friendships grows quadratically
with the number of people, since any pair of people can be friends.
Worse, simple concepts such as transitivity (friends of your friends
are likely to be your friends as well) lead to a cubic number of
ground formulae.  Thus, a domain of 1000 people has 1 million possible
friendships and 1 billion possible transitivity relations.  Standard
inference methods do not scale well to these sizes, especially when
formula weights are large, representing strong interactions among the
variables.  The opportunity is that these large models have a great
deal of symmetry, structure that can be exploited by ``lifted''
inference algorithms~(cf. \cite{braz&al05,singla_2008}).
Lifted inference works by grouping variables or variable
configurations together, leading to smaller, simpler inference
problems.  For example, when there is no evidence to differentiate
Anna and Bob, the probability that each smokes must be identical.
Depending on the MLN structure and evidence, lifted inference can
compute exact or approximate probabilities over millions of variables
without instantiating or reasoning about the entire network.  Another
technique for scaling inference is to focus on just the subset of the
network required to answer a particular query or set of
queries~\cite{riedel08}.
Related methods include lazy inference~\cite{singla_2006,poon_2008}, which only
instantiates the necessary part of the network, and coarse-to-fine
inference~\cite{kiddon&domingos11}, which uses more complex models to
refine the results of simpler models.

% OLD
%\eat{
%A key limiting factor for Markov logic inference and learning is the complexity of the partition function for the underlying graphical model. To address this challenge, a number of approaches for approximate inference have been developed, typically by combining logical methods with probabilistic ones. For example, MC-SAT \cite{poon_2006} combines SAT solvers with Markov chain Monte Carlo (MCMC) to effectively handle deterministic constraints in sampling. Both lazy inference \cite{singla_2006,poon_2008} and lifted inference \cite{singla_2008} leverage the compact first-order templates to avoid the combinatorial explosion in grounding the entire Markov network with high-order dependencies. 
%}

% NEW DBL TODO: Revision pass on this.  Keep high-level ideas clear,
% avoid excessive jargon.  Mention logistic regression connection?
% Other MN learning methods?  ...?
The parameter values in an MLN are critical to its effectiveness.  For
some applications, simply setting the relative weights of different
formulae is sufficient; for example, a domain expert may be able to
identify some formulae as hard constraints, others as strong
constraints, and others as weak constraints.  In many cases, however,
the best performance is obtained by carefully tuning these to fit
previously observed training data.  As with other probabilistic
models, MLN weight learning is based on maximizing the (penalized)
likelihood of the training data.  Given fully-observed training data,
this is a convex optimization problem that can be solved with gradient
descent or second-order optimization methods.  However, computing the
gradient requires running inference in the model to compute the
expected number of satisfied groundings of each formula.  Furthermore,
the gradient is often very ill-conditioned, since some formulae have
many more satisfying groundings than others.  The standard solution is
to approximate the gradient with a short run of Markov chain Monte
Carlo, similar to persistent contrastive divergence methods used to
train restricted Boltzmann machines~\cite{tieleman08}, and to
precondition the gradient by dividing by its variance in each
dimension~\cite{lowd&domingos07b}.  A popular alternative is
optimizing pseudolikelihood~\cite{besag75}, which does not require
running inference but may handle long chains of evidence poorly.

% DBL: List other applications, such as knowledge refinement,
% knowledge-aware ontology matching and knowledge translation?

\begin{figure}[tb]
\centering
\begin{subfigure}{.5\textwidth}
\centering
\includegraphics[width=.8\linewidth]{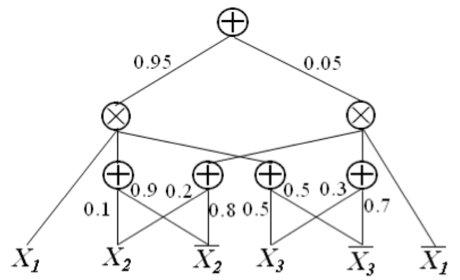}
\caption{}
\label{spn_junction_tree}
\end{subfigure}%
\begin{subfigure}{.5\textwidth}
\includegraphics[width=.8\linewidth]{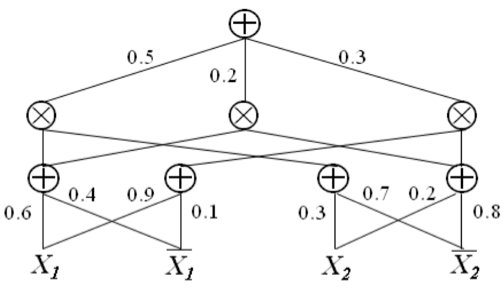}
\caption{}
\label{spn_naive_bayes}
\end{subfigure}
\caption{(a) Example SPNs implementing a junction tree with clusters $(X1 \text{ and } X2)$ and $(X1 \text{ and } X3)$ and separator $X1$. (b) Naive Bayes model with variables $X1$ and $X2$ and three components.}
\end{figure}

Most recently, a new paradigm has emerged by tightly integrating
learning with inference, as exemplified by sum-product networks (SPNs)
\cite{poon_2011}. The development of SPNs is motivated by seeking the
general conditions under which the partition function is tractable.
SPNs are directed acyclic graphs with variables as leaves, sums and
products as internal nodes, and weighted edges.
Figure~\ref{spn_junction_tree} shows an example SPNs implementing a
junction tree with clusters $(X1 \text{ and } X2)$ and $(X1 \text{ and }
X3)$ and separator $X1$, and Figure~\ref{spn_naive_bayes} depicts a
Naive Bayes model with variables $X1$ and $X2$ and three components.  

% DBL NEW: Added short inference tutorial to show more about how SPNs work.
To compute the probability of a partial or complete variable
configuration, set the values of the leaf variables and compute the
value of each sum and product node, starting at the bottom of the SPN.
For example, to compute $P(X_1=\mbox{true},X_3=\mbox{false})$ in the
SPN in Figure~\ref{spn_junction_tree}, set the value of
$\overline{X}_1$ and $X_3$ to 0, since these variable values contradict
the specified configuration, and set all other leaves to 1.  Each of
the four lower sum nodes is a weighted sum of two leaf values; from
left to right, they evaluate to $1$, $1$, $0.5$, and $0.3$.  The
product nodes above them evaluate to $0.5$ and $0.3$, and the overall
root equals $0.95 \cdot 0.5 + 0.05 \cdot 0.3 = 0.49$.  Thus, inference
follows the structure of the SPN, allowing arbitrary probability
queries to be answered in linear time.

It has been shown that if an SPN satisfies very general conditions called completeness and consistency, then it represents the partition function and all marginals of some graphical model. Essentially all tractable graphical models can be cast as SPNs, but SPNs are also strictly more general. SPNs achieve this by leveraging the key insight of dynamic programming, where intermediate results are captured (by internal sums and products) and reused. In this aspect, SPNs are similar to many existing formalisms such as $AND/OR$ graphs and arithmetic circuits. However, while these formalisms are normally used as compilation targets for graphical models, SPNs are a general probabilistic model class in its own right, with efficient learning methods developed based on backpropagation and EM. Experiments show that inference and learning with SPNs can be both faster and more accurate than with standard deep networks. For example, SPNs perform image completion better than many state-of-the-art deep networks for this task. SPNs also have intriguing potential connections to the architecture of the cortex (cf. \cite{poon_2011}).

% DBL NEW: RSPNS!
Relational sum-product networks (RSPNs) extend SPNs to handle
relational domains such as social networks and cellular
pathways~\cite{nath&domingos15}.  Like MLNs, an RSPN represents a
probability distribution over the attributes and relations among a set
of objects.  Unlike MLNs, inference in an RSPN is always tractable.
To achieve this, RSPNs define a set of object classes.  A class
specifies the possible attributes of each object, the classes of its
component parts, and the possible relations among the parts.  For
example, Figure~\ref{rspn} shows a partial class specification for a
simple political domain.  A `Region' may contain any number of nations, and
two nations within a region may be adjacent, in conflict, or both.  A
`Nation' contains a government object and any number of person objects,
each of which may or may not support the government.  A class also
specifies a tractable distribution over the attributes, parts, and
relations.  The structure and parameters of SPNs can be learned from
data.  RSPNs have been applied to social network and automated
debugging applications, where they were much faster and more accurate
than MLNs.
% DBL: I think we should mention something more about exchangeability,
% since that's key to RSPN efficiency, but I don't know how to work it
% in smoothly.

\begin{figure}
\begin{verbatim}
class Region:
  exchangeable part Nation
  relation Adjacent(Nation,Nation)
  relation Conflict(Nation,Nation)

class Nation:
  unique part Government
  exchangeable part Person
  attribute HighGDP
  relation Supports(Person,Government)
\end{verbatim}
\caption{Partial class definition for a relational sum-product network
in a simple political domain.}
\label{rspn}
\end{figure}

% DBL: Talk about structure learning of SPNs? (and briefly clarify
% parameter learning)

% DBL: Keep emphasis on integrating learning and reasoning --
% structure learning in SPNs is effectively learning a structure for
% reasoning over the (symbolic) inputs.

In summary, due to the achieved combination between symbolic
representations and subsymbolic (in this case, graphical) features and forms of reasoning, MLNs
and related formalisms have found a growing number of
applications and gained popularity in AI in general.  They can
therefore be seen as success stories of symbolic-numerical integration
closely related to the original domain of neurally-plausible or -inspired implementations of
high-level cognitive reasoning, exemplifying valuable theory work with
high relevance to computer science and, thus, complementing the aforementioned
examples---among others given by {\sc NSCA} and Penalty Logic---as relevant neural-symbolic applications
on the systems level.

\section{Relating Neural-Symbolic Systems to Human-Level Artificial Intelligence} \label{hlai}
Following the quite technical exposition of the last two sections, we now return to more conceptually-motivated considerations focusing on the relation between neural-symbolic integration and research in Human-Level Artificial Intelligence (HLAI), before identifying corresponding challenges and research opportunities.

HLAI, understood as the quest for artificially-created entities with human-like abilities, has been pursued by humanity since the invention of machines. It has also been a driving force in establishing artificial intelligence as a discipline in the 1950s. 20th-century AI, however, has developed into a much narrower direction, focussing more and more on special-purpose and single-method driven solutions for problems which were once (or still are) considered to be challenging, like game playing, problem solving, natural language understanding, computer vision, cognitive robotics, and many others. In our opinion, 20th-century AI can therefore be perceived as expert AI, producing and pursuing solutions for relevant specific tasks by using specialised methodologies designed for the tasks in question. We do not aim to say that this is a bad development---quite to the contrary, we think that this was (and still is) a very worthwhile and very successful endeavour with ample (and in some cases well-proven) scope for considerable impact on society. 

However---although having been one of the core ideas in the early days of AI---the pursuit of HLAI has been declining in the 20th century, presumably because the original vision of establishing systems with the desired capabilities turned out to be much harder to realise than expected in the beginning. Nevertheless, in recent years a rejuvenation of the original ideas has become apparent, driven on the one hand by the insight that certain complex tasks are outside the scope of specialised systems, and on the other hand by the discovery of new methodologies appropriate to address hard problems of general intelligence. Some examples for such new methodologies are the numerous approaches for machine learning, non-classical logic frameworks, or probabilistic reasoning, just to mention some of them. Furthermore, rapid developments in the neurosciences based on the invention of substantially refined means of recording and analysing neural activation patterns in the brain influenced this development. These are accompanied by interdisciplinary efforts within the cognitive science community, including psychologists and linguists with similar visions. 

It is apparent that the realisation of HLAI requires the cross-disciplinary integration of ideas, methods, and theories. Indeed we believe that disciplines such as (narrow) artificial intelligence, neuroscience, psychology, and computational linguistics will have to converge substantially before we can hope to realise human-like artificially intelligent systems. One of the central questions in this pursuit is thus a meta-question: What are concrete lines of research which can be pursued in the immediate future in order to advance in the right direction and to avoid progress towards---currently prominently discussed, albeit highly speculative---less desirable technological or social scenarios? The general vision does not give any answers to this, and while it is obvious that we require some grand all-encompassing interdisciplinary theories for HLAI, we cannot hope to achieve this in one giant leap. For practical purposes---out of pure necessity since we cannot, and for principled reasons should not even attempt to, shred our scientific inheritance \cite{besold_2013}---we require the identification of next steps, of particular topics which are narrow enough so that they can be pursued, but general enough so that they can advance us into the right direction. This brings us back to the topic of this survey article, neural-symbolic integration.

The proposal for neural-symbolic integration as such a research direction starts from two observations. These correspond to the two principled perspectives intersecting and combining in neural-symbolic integration:

(i) The physical implementation of our mind is based on the neural system, i.e. on a network of neurones as identified and investigated in the neurosciences. If we hope to achieve HLAI, we cannot ignore this neural or subsymbolic aspect of biological intelligent systems. 

(ii) Formal modelling of complex tasks and human---in particular, abstract---thinking is based on symbol manipulation, complex symbolic data structures (like graphs, trees, shapes, and grammars) and symbolic logic. At present, there exists no viable alternative to symbolic approaches in order to encode complex tasks. 

These two perspectives however---the neural and the symbolic---are substantially orthogonal to each other in terms of the state-of-the-art in the corresponding disciplines. As also evidenced by the previous sections, symbolically understanding neural systems is hard and requires significant amounts of refined theory and engineering while still falling short of providing general solutions. Furthermore, it is quite unclear at present how symbolic processing at large emerges from neural activity of complex neural systems. Therefore, symbolic knowledge representation and the manipulation of complex data structures used for representing knowledge at the level required for HLAI is way outside the scope of current artificial neural approaches. 

At the same time humans---using their neural wetware, i.e. their brains---are able to deal successfully with symbolic tasks, to manipulate symbolic formalisms, to represent knowledge using them, and to solve complex problems based on them. In total, there is a considerable mismatch between human neurophysiology and cognitive capabilities as role models for HLAI on the one hand, and theories and computational models for neural systems and symbolic processing on the other hand. 

It is our belief that significant progress in HLAI requires the reconciliation of neural and symbolic approaches in terms of theories and computational models. We believe that this reconciliation is as much central for the advancement of HLAI as it is for cognitive science and cognitive neuroscience (as described in Section \ref{mental_models}). We also believe that the pursuit of this reconciliation is timely and feasible based on the current state of the art, and will continue by briefly mentioning some recent developments in neural-symbolic integration which we consider to be of particular importance for the HLAI discussion. Further pointers can be found, e.g., in \cite{bader_2005,hammer_2007,garcez_2009}.
 
The line of investigation we want to mention takes its starting point from computational models in (narrow) AI and machine learning. It attempts to formally define and practically implement systems based on ANNs which are capable of learning and dealing with logic representations of complex domains and inferences on these representations.  While this can be traced back to \cite{mcculloch_1943} 's landmark paper on the relation between propositional logic and binary threshold ANNs, it has been largely dormant until the 1990s, where (as already mentioned in previous sections) first neural-symbolic learning systems based on these ideas were realised---cf., among others, \cite{towell_1994,garcez_1999,garcez_2002}. While these initial systems were still confined to propositional logics, in recent years systems with similar capabilities based on first-order logic have been realised---cf., for instance, \cite{gust_2007} or Section \ref{connectionist_first-order_logic}. It is to be noted, however, that these systems---despite the fact that they provide a conceptual breakthrough in symbol processing by ANNs---are still severely limited in their scope and applicability, and improvements in these directions do not appear to be straightforward at all. In order to overcome such problems, we also require new ideas borrowed from other disciplines, in order to establish neural-symbolic systems which are driven by the HLAI vision. Results from cognitive psychology on particularities of human thinking which are usually not covered by standard logical methods need to be included. Recent paradigms for ANNs which are more strongly inspired from neuroscience---as discussed, for example, in \cite{maass_2002}---need to be investigated for neural-symbolic integration. On top of this, we require creative new ideas borrowed, among others, from dynamical systems theory or organic computing to further the topic. 

The presented selection is considered to be exemplary and there are several other efforts which could be mentioned. However, we selected this line of research for presentation in the last paragraph as to us it appears to be typical and representative in that it is driven by computer science, machine learning, and classical AI research augmented with ideas from cognitive science and neuroscience.

\section{(Most) Recent Developments and Work in Progress from the Neural-Symbolic Neighbourhood} \label{recent_developments}
Before concluding this opinionated survey with two sections dedicated to future challenges and directions (Section \ref{future_directions}) and to some short overarching remarks (Section \ref{conclusion}), in the following we will highlight recent work surfacing outside the traditional core field but in the topical proximity of neural-symbolic integration with high relevance and potential impact for the corresponding core questions.

\subsection{Other Paradigms in Computation and Representation Narrowing the Neural-Symbolic Gap}
We first want to direct the reader's attention to a recently proposed neuro-computational mechanism called ``conceptors'' \cite{jaeger_conceptors_2014}. Broadly speaking, conceptors are a proposed mathematical, computational, and neural model combining two basic ideas, namely that the processing modes of a recurrent neural network (RNN) can be characterised by the geometries of the associated state clouds, and that the associated processing mode is selected and stabilised if the states of the respective RNN are filtered to remain within a certain corresponding state cloud. When taken together, this basically amounts to a possibility to control a multiplicity of processing modes of one single RNN, introducing a form of top-down control to the bottom-up connectionist network. While the underlying mathematical considerations are fairly elaborate, the overall approach can be conceptualised as a two-step approach. First, the different elipsoids envelopping the differently shaped state clouds a driven RNN exhibits when exposed to different dynamical input patterns each give a different conceptor. Once the driving patterns have been stored in the network (i.e., the network has learned to replicate the pattern-driven state sequences in the absence of the driver stimuli, performing a type of self-simulation), they can be selected and stably re-generated by inserting the corresponding conceptor filters (i.e., the geometry of the elipsoid serves as decision mechanism of the filter) in the network's update loop. This partially geometric nature of the approach also offers another advantage in that conceptors can be combined by operations akin (and partially co-extensional) to Boolean logic, thus introducing an additional dimension of semantic interpretation especially catering to the symbolic side of neural-symbolic integration.

\begin{figure}
\centering
\includegraphics[width = 0.56125\textwidth]{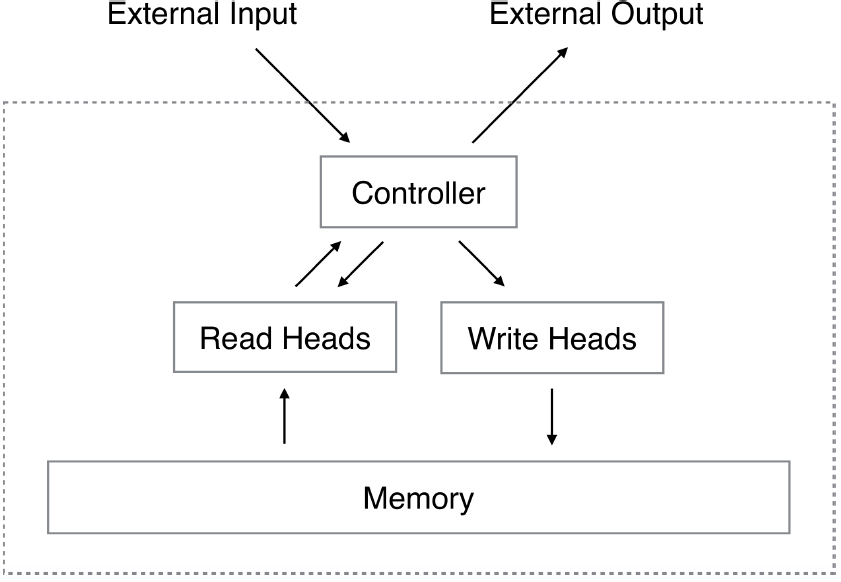}
\caption{The basic structure of a Neural Turing Machine (taken from \cite{graves_2014}): The controller network receives external inputs and emits outputs during each update cycle. Simultaneously, it reads to and writes from a memory matrix via parallel read and write heads.\label{NTM}}
\end{figure}

While conceptors are an approach to equipping sub-symbolic models in form of RNNs with certain symbolic capacities and properties, the second recent research result we want to mention aims at a connectionist implementation of Von Neumann's computing architecture \cite{von_neumann_1945} as the prototypical symbolic computer model. ``Neural Turing Machines'' (NTMs), introduced in \cite{graves_2014}, basically couple RNNs with a large, addressable memory yielding the end-to-end differentiable and gradient-decent trainable analog of Turing's memory-tape enriched finite-state machines. In terms of architecture, an NTM is constituted by a neural network controller and a memory bank (see Figure \ref{NTM}). The controller manages the interaction with the environment via input and output vectors and interacts with a memory matrix using selective read and write operations. The main feature of NTMs is their end-to-end differentiability which allows for training with gradient-descent approaches. In order to introduce this capacity, read and write operations have been defined in a way which allows them to interact to different degrees with all elements in the memory (modulated by a focus mechanism limiting the actual range of interaction), resulting in a probabilistic mode of reading and writing (different from the highly specific discrete read/write operations a Turing machine performs). Also, interaction with the memory tends to be highly sparse, biasing the model towards interference-free data storage. Which memory location to attend to is determined by specialised network outputs parameterising the read and write operations over the memory matrix, defining a normalised weighting over the rows of the matrix. These network outputs thereby can focus attention to a sharply defined single memory location or spread it out to weakly cover the memory at multiple adjacent locations. Several experiments also reported in \cite{graves_2014} demonstrate that NTMs are generally capable of learning simple algorithms from example data and subsequently generalise these algorithms outside of the training regime.

Another approach to combining the learning capacities of connectionist networks with a form of read- and writeable memory, as well as inference capacities, has been presented in the form of so called ``memory networks'' \cite{weston_2015}. A memory network consists of an array of indexed objects (such as arrays or vectors) $m$ constituting the memory component, an input feature map $I$ converting incoming input to the respective internal feature representation, a generalisation $G$ updating (and/or possibly compressing and generalising) old memories given the new input, an output feature map $O$ producing a new output given the new input and the current memory state, and finally a response $R$ converting the output into the final format. Except for $m$, the remaining four components $I$, $G$, $O$, and $R$ could potentially be learned. Against this backdrop, if an input $x$ arrives at the memory network, $x$ is converted to an internal feature representation $I(x)$ before updating the memories $m_i$ within $m$ given $x$: $\forall i: m_i = G(m_i, I(x), m)$. Subsequently, output features $o = O(I(x),m)$ are computed given the new input and the memory, and the output features are decoded into a final response $r = R(O)$. The flow of the model remains the same during both training and testing, with the only difference being that the model parameters $I$, $G$, $O$, and $R$ stay unchanged at test time. Due to this very general characterisation, memory networks can be flexible concerning the possible implementation mechanisms for $I$, $G$, $O$, and $R$, generally allowing the use of standard models from machine learning (such as RNNs, SVMs, or decision trees). In \cite{weston_2015} also an account of an instantiation of the general model is given, describing an implementation of a memory network for application in a text-based question answering (QA) setting using neural networks as components. The resulting system is then applied and evaluated separately in a large-scale QA and a simulated QA setting (the latter also including some QA with previously unseen words), before combining both setups in a final hybrid large-scale and simulated QA system serving as proof of concept for the possible future capacities of the approach.

\subsection{Other Application Systems Narrowing the Neural-Symbolic Gap}
Following the previous discussion of neural-symbolically relevant advances on the architectural and methodological level, in this subsection we will focus on several recent application systems which are partially or fully based on connectionist approaches but are successfully applied to tasks of predominantly symbolic nature.

The first application area which recently has seen rapid development and systems operating on a previously unachieved, qualitatively new level of performance is the domain of vision-based tasks such as the semantic labelling of images according to their pictorial content. In \cite{vinyals_2014} and \cite{karpathy_2015} two approaches to related problems have been presented. While the generative, deep RNN model in \cite{vinyals_2014} generates natural language sentences describing the content of input images, \cite{karpathy_2015} introduces an approach to recognition and labelling tasks for the content of different image regions (corresponding to different objects and their properties in the pictures). The model from \cite{vinyals_2014} combines an input convolutional neural network, used for encoding the image (the network is pre-trained for an image classification task, and the last hidden layer serves as interface to the following natural language decoder), with an output RNN, generating natural language sentences based on the processed data of the input network. The result is a fully stochastic gradient descent-trainable neural network combining, in its sub-networks a vision and a language model, and---when put in application---significantly outperforming previous approaches and narrowing the gap to human performance on the used test sets. In \cite{karpathy_2015}, a region convolutional neural network is applied to the identification of objects (and the corresponding regions) in images and their subsequent encoding in a multimodal embedding space, together with a bidirectional RNN which computes word representations from input sentences and also embeds them in the same space. Once this mapping of images and sentences into the embedding space has been achieved, an alignment objective in terms of an image-sentence score (a function of the individual region-word scores) is introduced, tying together both parts of the model. The resulting system again performs better than previous approaches, showing state of the art performance in image-sentence ranking benchmarks and outperforming previous retrieval baselines in fullframe and region-level experiments. Both models and the corresponding implementation systems are remarkable in that classically tasks involving semantic descriptions had been associated with databases containing some form of background knowledge, and the relevant picture processing/computer vision approaches had for a long time also relied on rule-based techniques. The described work gives proof that this type of task---which clearly conceptually still operates on a symbolic level---generally can also be addressed in practice in purely connectionist architectures.

Another closely related type of task is visual analogy-making as, for instance, exemplified in transforming a query image according to an example pair of related images. In \cite{reed_2015}, the authors describe a way how to tackle this challenge again applying deep convolutional neural networks. In their approach a first network learns to operate as a deep encoder function mapping images to an embedding space (which due to its internal structure allows to perform some forms of reasoning about analogies via fairly standard mathematical operations), before a second network serves as a deep decoder function mapping from the embedding back to the image space. The resulting models are capable of learning analogy-making based on appearance, rotation, 3D pose, and several object attributes, covering a variety of tasks ranging from analogy completion, to animation transfer between video game characters, and 3D pose transfer and rotation. Also here, in general analogy models previously have mostly been symbol-based, and generative models of images had to encode significant prior background knowledge and restrict the allowed transformations, making the general task domain mostly symbolic in nature.

A different domain is addressed by the work described in \cite{foerster_2016}. There, a deep distributed recurrent Q-network is successfully applied to learning communication protocols for teams of agents in coordination tasks in a partially observable multi-agent setting. This type of task requires agents to coordinate their behaviour maximising their common payoff while dealing with uncertainty about both, the hidden state of the environment and the information state and future actions of the members of their team. In order to solve this challenge, so called ``deep distributed recurrent Q-networks'' (DDRQN) have been introdcued. These networks are based on independent deep Q-network agents with Long Short-Term Memory (LSTM) networks \cite{hochreiter_1997}---a currently very popular type of RNN for tasks involving the processing of sequential data---augmented by (i) supplying each agent with its previous action as input on the following time step (enabling agents to approximate their own action-observation history), (ii) sharing a single network's weights between all agents and conditioning on the agent's ID (enabling fast learning and diverse behaviour), and (iii) disabling experience replay (avoiding problems with non-stationarity arising from the simultaneous learning of multiple agents). The resulting system proves capable of solving (for limited numbers of agents) two classical logic riddles necessarily requiring collaborative communication and information exchange between players in their optimal solution strategies. While future work still will have to prove the scalability of the approach, the work in \cite{foerster_2016} nonetheless is a proof of concept for the ability of connectionist networks to learn communication strategies in multi-agent scenarios---which is another instance of a genuinely symbolic capacity that has successfully been implemented in a neural network architecture.

The final example of a successful application system addressing aspects of neural-symbolic integration which we want to mention is the AlphaGo Go-playing system presented in \cite{silver_2016}. While Go had for a long time proven to be computationally unfeasible to solve due to the quickly growing game tree and corresponding search space, the authors introduce an architecture equipped with deep neural network-based move selection and position evaluation functions (learned using a combination of supervised learning from human expert games and reinforcement learning from massively parallel self-play). Integrating these, in the form of policy and value networks, with a Monte Carlo tree search lookahead approach, the resulting high-performance tree search engine allows to efficiently explore the game tree and play the game at a level on par with the world's best human players. Different from the previous examples from the visual domain, the resulting architecture is closer to being genuinely neural-symbolic in its combination between trained connectionist networks for pruning and narrowing down the search space, and the subsequent Monte Carlo algorithm selecting actions by lookahead tree search. With its previously unachieved performance AlphaGo constitutes a prime example of the possibilities introduced by combining connectionist and symbolic approaches and techniques.

\subsection{Other Research Efforts and Programs Narrowing the Neural-Symbolic Gap}
In this final subsection we want to focus on three different lines of work which stand out among the usual work in neural-symbolic integration in that they either start out from a different perspective than neural computation or logic-based modelling, or apply techniques and approaches which are not commonly used in the field. Examples for such ``non-classical neural-symbolic research programs'' can be found, for instance, in certain investigations into methods and mechanisms of multi-agent systems, in empirical attempts at understanding RNNs by using language models as testbed providing data for qualitative analyses, and in theoretical computer science and complexity theory projects targeting the practical differences between symbolic and connectionist approaches with theoretical means. We describe three recent suggestions as prototypical cases in point. 

\textbf{(i) Anchoring Knowledge in Interaction in Multi-Agent Systems:} The typical setting in multi-agent systems encompasses an agent exploring the environment, learning from it, reasoning about its internal state given the input, acting and, in doing so, changing the environment, exploring further, and so on as part of a permanent cycle \cite{cutsuridis_2011}. One of the biggest challenges in this context is the (bidirectional) integration between low-level sensing and interacting as well as the formation of high-level knowledge and subsequent reasoning. In \cite{besold_et_al_2015} an interdisciplinary group of researchers spanning from cognitive psychology through robotics to formal ontology repair and reasoning sketch conceptually (what amounts to) a strongly cognitively-motivated neural-symbolic architecture and model of a situated agent's knowledge acquisition through interaction with the environment in a permanent cycle of learning through experience, higher-order deliberation, and theory formation and revision.

In the corresponding research program---further introduced and actually framed in terms of neural-symbolic integration in \cite{besold_kuehnberger_2015}---different topics also mentioned in the present survey article are combined for that purpose. Starting out from computational neuroscience and network-level cognitive modelling (as represented, e.g., by the conceptors framework described above) combined with psychological considerations on embodied interaction as part of knowledge formation, low-level representations of the agent's sensing are created as output of the ground layer of the envisioned architecture/model. This output then is fed into a second layer performing an extended form of anchoring \cite{coradeschi_2000} not only grounding symbols referring to perceived physical objects but also dynamically adapting and repairing acquired mappings between environment and internal representation. The enhanced low-level representations as output of the anchoring layer are then in turn fed to the---in terms of neural-symbolic integration most classical---lifting layer, i.e., to an architectural module repeatedly applying methods combining neural learning and temporal knowledge representation in stochastic networks (as, e.g., already discussed in the form of RBMs in the context of the {\sc NSCA} framework in Section \ref{learning_and_reasoning}) to continuously further increase the abstraction level of the processed information towards a high-level symbolic representation thereof. Once a sufficiently general level has been reached, a final layer combining elaborate cognitively-inspired reasoning methods on the symbolic level (such as, for instance, analogy-making, concept blending, and concept reformation) then is used for maintaining a model of the agent and the world on the level of explicit declarative knowledge \cite{bundy_2013}, which nonetheless is intended to feed back into the lower layers---among others by serving as guidance for the augmented anchoring process.

\textbf{(ii) Visualising and Understanding Recurrent Networks:} LSTMs stand out between current deep network models due to their capacity to store and retrieve information over prolonged time periods using built-in constant error carousels governed by explicit gating mechanisms. Still, while having been used in several high-profile application studies, this far only little is known about the precise properties and representation capacities of LSTMs. In \cite{karpathy_2015_2}, a first step towards clarifying some of the relevant questions is described and results are presented which are also relevant from a neural-symbolic perspective. The authors empirically shed light on aspects of representation, predictions and error types in LSTM networks, uncover the existence of interpretable cells within the network which keep track of long-range dependencies in the respective input data, and suggest that the remarkable performance of LSTMs might be due to long-range  structural dependencies.

In order to gain insight into the inner workings of LSTM networks, character-level language models are used as an interpretable testbed for several computational experiments. In theory, by making use of its memory cells an LSTM should be capable to remember long-range information and keep track of different attributes of the respective input it is just processing. Also, using text processing as an example, manually setting up a memory cell in such a way that it keeps track of whether it is inside a quoted string or not is not overly challenging. Still, whether LSTMs actually resort to this type of internal structure (i.e., forming interpretable cells with dedicated, more abstract tasks) was not as clear. The empirical findings in \cite{karpathy_2015_2} indicate  that such ``task cells'' indeed exist, in the case of processing program code as textual input ranging from cells checking for parenthesis after an if statement to cells acting as line length counter or tracking the indentation of blocks of code. In a second step, using finite horizon $n$-gram models as comparandum, it is shown that LSTM networks are remarkably good at keeping track of long-range interactions and dependencies within the input data. 

From the neural-symbolic point of view, the work reported in \cite{karpathy_2015_2} is important in at least two ways. On the one hand, the findings relating to the existence of interpretable long-range interactions and associated cells in the network setup actually take LSTMs one step closer to the level of symbolic interpretability and away from the completely distributed and non-localised classical connectionist setting. On the other hand, the overall approach and the set of methods applied in the corresponding empirical studies and analyses (including qualitative visualisation experiments, statistical evaluations of cell activations, and the comparison to the $n$-gram models) suggest potential techniques which could be used in more detailed follow-up studies about knowledge extraction from LSTMs and similar assessments of other types of recurrent neural networks in general.

\textbf{(iii) Identifying and Exploring Differences in Complexity:} Contrasting the multi-agent approach described in (i) and the empirical analysis of LSTM networks in (ii), the second research program proposed in \cite{besold_kuehnberger_2015} approaches the differences between ANNs and logics from the angle of theoretical computer science and complexity theory. In the suggested line of work recent advances in the modelling and analysis of connectionist networks and new developments and tools for investigating previously unconsidered properties of symbolic formalisms shall be combined in an attempt at providing an explanation for the empirical differences in terms of applicability and performance between neural and logical approaches very quickly showing up in application scenarios (which seem to harshly contradict the formal equivalence between both described in Section \ref{prolegomena}).

To this end, it is suggested that the form and nature of the the polynomial overhead as computational-complexity difference between paradigms should be examined in more detail. The underlying hope is that a closer look at this overhead might shed light on previously unconsidered factors as hitherto most complexity results have been established using exclusively $TIME$ and $SPACE$ as classical resources for this form of analysis, and most analyses have left out more precise investigations of the remaining polynomial overhead after establishing tractability. Against this background, the working hypotheses for the program are that for a fully informative analysis other resources might have to be taken into account on the connectionist side (such as the number of spikes a spiking network requires during computation, or the number of samples needed for convergence from an initial network state to the stationary distribution) and that additional formal tools and techniques will have to be applied on the symbolic side (such as parameterised-complexity methods which take into account problem- and application-specific properties of problem classes, or descriptive-complexity theory which also considers the polynomial-time and the logarithmic-time hierarchy).

The results of these efforts promise to contribute to resolving some of the basic theoretical and practical tensions arising when comparing the suitability and feasibility of an application of certain formalisms to different types of tasks and domains. Which, in turn, then is expected to also successively provide information on general challenges (and their potential remedies) which would have to be overcome for advancing towards neural-symbolic integration on the formal and theoretical side.

\section{Challenges and Future Directions} \label{future_directions}
In this second-to-last section of our survey we give an overview of what we consider the currently most pressing and/or---if addressed successfully---potentially most rewarding theoretical and practical questions and topics on the way towards bridging the gap between connectionist computation and high-level reasoning.

As described throughout this article, if connectionism is to be an alternative paradigm to logic-based artificial intelligence, ANNs must be able to compute symbolic reasoning efficiently and effectively, combining the fault-tolerance of the connectionist component with the ``brittleness and rigidity'' of its symbolic counterpart. On the more architecturally-oriented side, by integrating connectionist and symbolic systems using some of the previously sketched or similar approaches, hybrid architectures \cite{sun_2002} seek to tackle this problem and offer a principled way of computing and learning with knowledge of various types and on different levels of abstraction. Starting out from multi-modular approaches such as, for instance, the {\sc CLARION} architecture \cite{sun_2003} in cognitive modelling or the vision of symbolic/sub-symbolic multi-context systems \cite{besold_2010} in data integration and reasoning, they have the potential to open up the way and serve as foundation for the convergence between the paradigms and the development of fully-integrated monolithic neural-symbolic systems. Nonetheless, the development of such systems is currently still in its infancy and often is motivated much more by concrete engineering imperatives on a case-by-case basis than by coordinated and targeted research efforts aiming at testing general approaches and understanding the general mechanisms and constraints at work.

On a more positive note, from a more formally-centred perspective it is possible to identify several current points of approximation and potential confluence: deep belief networks are based on Boltzmann machines as an early symmetric ANN model, and related to Bayesian networks. Certain neural-symbolic recurrent networks are very similar to dynamic Bayesian networks. Connectionist modal logic uses modal logic as a language for reasoning about uncertainty in a connectionist framework. Markov logic networks combine logical symbols and probability. In summary it can, thus, be observed that the objectives indeed seem to be converging. Still, also here the current state of affairs is not yet satisfactory as too often too little consideration is given to an integration on the level of mechanisms, allowing them to remain varied instead of triggering convergence also along this (potentially even more relevant) dimension.

In general, many questions and limits to integration hitherto are unsolved and sometimes even unaddressed. Currently, the focus still predominantly lies on either learning or reasoning, with efforts aiming to improve one using state-of-the-art models from the other, rather than truly integrating both in a principled way with contribution to both areas. In addition to, for instance, \cite{garcez_2015} we now turn to some of the (many) remaining challenges: we have seen that neural-symbolic systems are composed of (i) translations from logic to network, (ii) machine learning and reasoning, (iii) translation from network to logic. Among the main challenges are in (i) finding the limits of representation, in (ii) finding representations that are amenable to integrated learning and reasoning, and in (iii) producing effective knowledge extraction from very large networks. 

What follows is a list of research issues related to challenges (i) to (iii) each of which could serve as basis for a research program in its own right:
\begin{itemize}
\item Reconciling first-order logic learning and first-order logic reasoning. 
\item Embracing semi-supervised and incremental learning.
\item Evaluating and analysing large-scale gains of massive parallelism. 
\item Implementing learning-reasoning-acting cycles in cognitive agents. 
\item Developing representations for learning which learn the ensemble structure. 
\item Rule extraction and interpretability for networks with thousands of neurones. 
\item Applying fibring in practice and actually learning the fibring function. 
\item Theoretically understanding the differences in application behaviour between connectionist and symbolic methods. 
\item Developing a proof theory and type theory for ANNs. 
\item Developing analogical and abductive neuro-symbolism with the aim of automating the scientific method. 
\item Investigating neural-symbolic models of attention focus and modelling and contrasting emotions and utility functions.
\end{itemize}

While some of the former recently have started to attract active attention from researchers---examples are the work on neural-symbolic multi-agent systems, the proposal for a research program addressing the learning-reasoning-acting cycle, or the effort to better theoretically understand the theoretical basis of application differences between neural and logical approaches outlined in Section \ref{recent_developments}---our knowledge about these issues is only limited and many questions still have to be asked and answered.

Also in terms of approach, several remarks seem in place: while seeing it as highly relevant concerning motivations and as source of inspiration, we agree that the direct bottom-up approach (i.e., brain modelling) is not productive given the current state of the art. Also, we agree that an engineering approach is valid but specific and will not answer the big questions asked, for instance, in more cognitively- or HLAI-oriented lines of research (including some of the bullet points above). Therefore, a foundational approach is needed combining all the different directions touched upon in this survey article, i.e., encompassing (at least) different elements of computer science (in particular, knowledge representation and machine learning), cognitive science, and cognitive neuroscience. 

Finally, in terms of applications, the following can be counted among the potential ``killer applications'' for neural-symbolic systems: ANNs have been very effective at image processing and feature extraction as evidenced, for instance, by \cite{ji_2013} and the work reported in the previous section. Yet, large-scale symbolic systems have been the norm for text processing (as most approaches currently in use rely on large ontologies, are inefficient to train from scratch, and heavily dependent on data pre-processing). Even if networks could be trained without providing starting knowledge to perform as well as, for example, WordNet, the networks would be very difficult to maintain and validate. Neural-symbolic systems capable of combining network models for image processing and legacy symbolic systems for text processing seem therefore ideally suited for multimodal processing. Here concepts such as fibring (our running example from Section \ref{prolegomena}) seem important so that symbolic systems and network models can be integrated loosely at the functional level. In this process, inconsistencies may arise, making the resolution of clashes a key issue. One approach is to see inconsistency as a trigger for learning, with new information in either part of the combined system serving to adjudicate the conflict (similar to the mechanisms envisioned within the learning-reasoning-acting framework described in the previous section). The immediate impact of this application would be considerable in many areas including the web, intelligent applications and tools, and security.

\section{Concluding Remarks} \label{conclusion}
This survey summarises initial links between the fields of computer science (and more specifically AI), cognitive science, cognitive neuroscience, and neural-symbolic computation. The links established are aimed at answering a long-standing dichotomy in the study of human and artificial intelligence, namely the perceived dichotomy of brain and mind, characterised within computer science by the symbolic approach to automated reasoning and the statistical approach of machine learning. 

We have discussed the main characteristics and some challenges for neural-symbolic integration. In a nutshell:
$$\text{neural-symbolic systems} = \text{connectionist machine} + \text{logical abstractions}$$

The need for rich, symbol-based knowledge representation formalisms to be incorporated into learning systems has explicitly been argued at least since \cite{valiant_1984}'s seminal paper, and support for combining logic and connectionism, or logic and learning was already one of Turing's own research endeavours \cite{pereira_2012}. 

Both the symbolic and connectionist paradigms embrace the approximate nature of human reasoning, but when taken individually have different virtues and deficiencies. Research into the integration of the two has important implications that can benefit computing, cognitive science, and even cognitive neuroscience. The limits of effective integration can be pursued through neural-symbolic methods, following the needs of different applications where the results of principled integration must be shown advantageous in practice in comparison with purely symbolic or purely connectionist systems. 

The challenges for neural-symbolic integration today emerge from the goal of effective integration, expressive reasoning, and robust learning. Computationally, there are challenges associated with the more practical aspects of the application of neural-symbolic systems in areas such as engineering, robotics, semantic web, etc. These challenges include the effective computation of logical models, the efficient extraction of comprehensible knowledge and, ultimately, striking of the right balance between tractability and expressiveness.

In summary, by paying attention to the developments on either side of the division between the symbolic and the sub-symbolic paradigms, we are getting closer to a unifying theory, or at least promoting a faster and principled development of cognitive and computing sciences and AI. To us this is the ultimate goal of neural-symbolic integration together with the associated provision of neural-symbolic systems with expressive reasoning and robust learning capabilities.

%\acks{Lorem ipsum...}

\vskip 0.2in
\bibliography{neural-symbolic_survey_arxiv}
\bibliographystyle{theapa}

\end{document}